\documentclass{article}
\usepackage{amssymb}

\usepackage[preprint]{corl_2025} % Uncomment for pre-prints (e.g., arxiv); This is like ``final'', but will remove the CORL footnote.

\usepackage{times}

\usepackage{graphics} % for pdf, bitmapped graphics files
\usepackage{amsmath} % assumes amsmath package installed
\usepackage{amssymb}  % assumes amsmath package installed
\usepackage{siunitx}
\usepackage{mathtools}
\usepackage{amsthm}
\usepackage{booktabs}
\usepackage{multirow}
\usepackage{bbm}
\usepackage[dvipsnames]{xcolor}

\usepackage{subcaption} % apparently overrides IEEE captions
\usepackage{enumerate}
\usepackage{mathrsfs}
\usepackage{svg}

\usepackage{graphicx}
\usepackage{adjustbox}
\usepackage{pbox}
\usepackage{dblfloatfix}
\usepackage{graphicx}
\usepackage{hyperref}
\usepackage{caption}
\usepackage[nospace]{cite}
\usepackage{microtype}

\usepackage{wrapfig}
\usepackage[ruled]{algorithm2e} % http://www.ctan.org/pkg/algorithm2e
\usepackage{xcolor} % http://www.ctan.org/tex-archive/macros/latex/contrib/xcolor
\usepackage{tikz}

\definecolor{mgreen}{RGB}{1,130,52}
\definecolor{mred}{RGB}{146,32,51}

\newcommand{\green}[1]{\textcolor{mgreen}{#1}}
\newcommand{\red}[1]{\textcolor{mred}{#1}}
\usetikzlibrary{fit,calc}
\colorlet{mypink}{red!40}
\colorlet{myblue}{cyan!60}

\colorlet{mypink}{red!40}
\colorlet{myblue}{cyan!60}

\usepackage{graphicx}
\usepackage{array,multirow,graphicx}
\usepackage{makecell}
\usepackage[T1]{fontenc}
\usepackage{microtype}
\usepackage{graphicx}
\usepackage{tikz}
\usetikzlibrary{positioning}
\usepackage{pgfplots}
\usepackage{xspace}
\usepackage[numbers]{natbib}
\usepackage{multicol}

\usepackage{amssymb}% http://ctan.org/pkg/amssymb
\usepackage{pifont}% http://ctan.org/pkg/pifont
\usepackage{soul}

\newcommand{\changed}[1]{\textcolor{black}{#1}}
 \newcommand{\removed}[1]{}
\newcommand{\OurMethod}{GraspQP}

\newcommand{\oursubsubsection}[1]{\par\noindent\textbf{#1}:}

\newcommand{\PSpan}[1]{
\texttt{pos}({#1})
}

\newtheorem{definition}{Definition}[section]
\newtheorem{theorem}{Theorem}[section]

\title{
\OurMethod: Differentiable Optimization of Force Closure for Diverse and Robust Dexterous Grasping
}

\author{
  René Zurbrügg \\
 ETH Zürich and ETH AI Center\\
  \texttt{zrene@ethz.ch} \\
  \And
  Andrei Cramariuc \\
 ETH Zürich \\
  \texttt{crandrei@ethz.ch} \\
  \And
  Marco Hutter \\
 ETH Zürich \\
  \texttt{mahutter@ethz.ch} \\
}

\begin{document}
\maketitle
% TODO REMVOE THIS ONCE SUBMITTING (PAGE NUMBERS)
% \thispagestyle{plain}
%\pagestyle{plain}

\begin{figure}[h]
    \centering
    \includegraphics[width=0.9\linewidth]{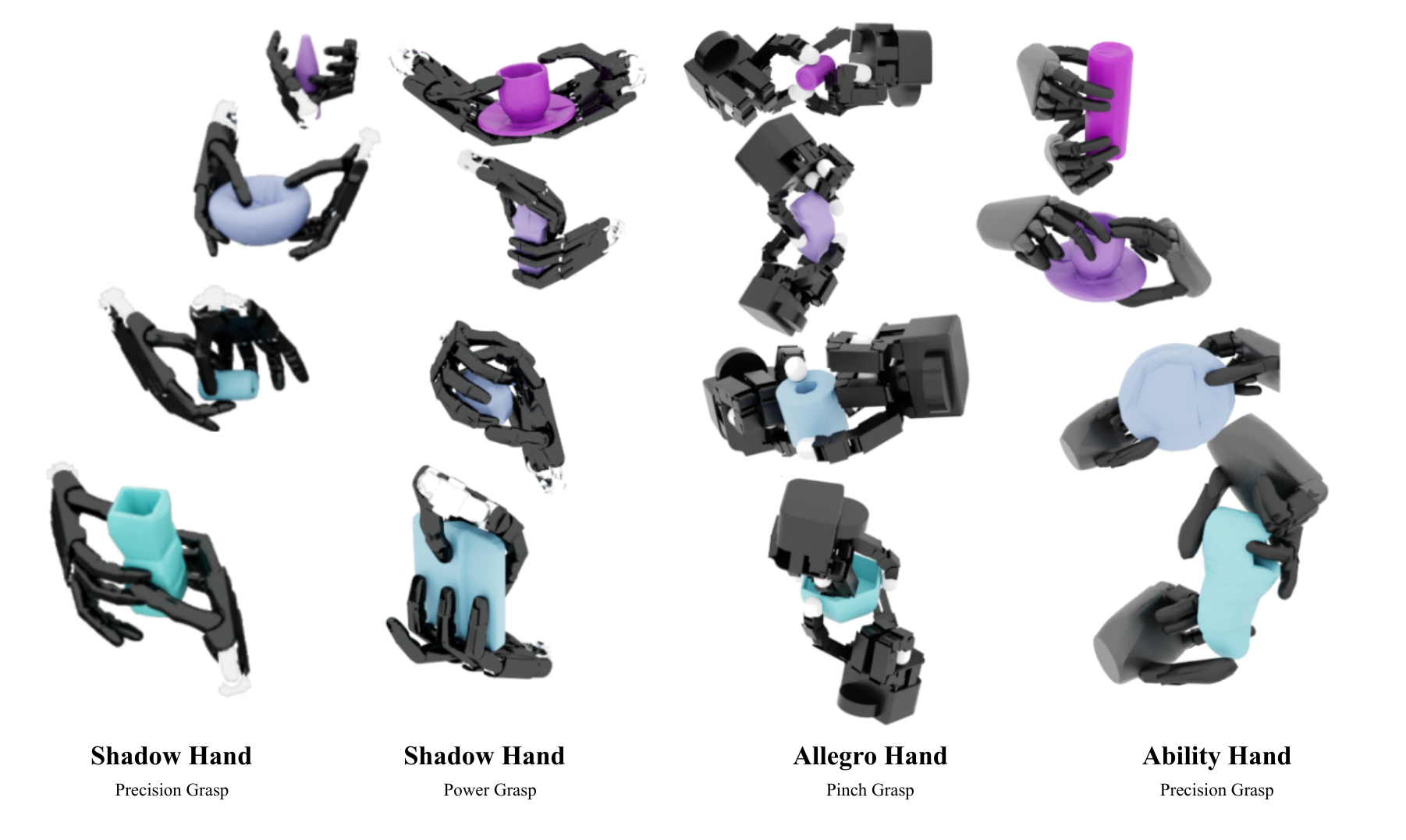}
    \caption{Visualization of predicted grasp poses for dexterous hands. % From left to right, shadow-hand, allegro and ability hand, grasps types raning from precision, pinch , precision and power grasps.
    }
    \label{fig:title-fig}
\end{figure}

\begin{abstract}
    Dexterous robotic hands enable versatile interactions due to the flexibility and adaptability of multi-fingered designs, allowing for a wide range of task-specific grasp configurations in diverse environments.
However, to fully exploit the capabilities of dexterous hands, access to diverse and high-quality grasp data is essential- whether for developing grasp prediction models from point clouds, training manipulation policies, or supporting high-level task planning with broader action options.
Existing approaches for dataset generation typically rely on sampling-based algorithms or simplified force-closure analysis, which tend to converge to power grasps and often exhibit limited diversity.
In this work, we propose a method to synthesize large-scale, diverse, and physically feasible grasps that extend beyond simple power grasps to include refined manipulations, such as pinches and tri-finger precision grasps.
We introduce a rigorous, differentiable energy formulation of force closure, implicitly defined through a Quadratic Program (QP).
Additionally, we present an adjusted optimization method (MALA*) that improves performance by dynamically rejecting gradient steps based on the distribution of energy values across all samples.
We extensively evaluate our approach and demonstrate significant improvements in both grasp diversity and the stability of final grasp predictions.
Finally, we provide a new, large-scale grasp dataset for 5,700 objects from DexGraspNet, comprising five different grippers and three distinct grasp types.\\
\emph{Dataset and Code}: \url{https://graspqp.github.io/}
\end{abstract}
    
\newpage

\section{Introduction}

Dexterous grasping is a fundamental capability in robotics, enabling a wide range of applications, including industrial automation, service robotics, and human-robot collaboration.
It is critical for allowing robots to interact with and manipulate objects in complex, unstructured environments, and remains a key focus of current research.
Central to progress in dexterous grasping is the ability to predict stable grasps for various morphologies and under partial observation. A task that heavily relies on learning from large-scale datasets~\citep{wang2023dexgraspnet, casas2024multigrippergraspdatasetroboticgrasping, turpin2023fastgraspd,9560844,columbia_database}.
However, existing datasets are costly to generate and often limited in their diversity of gripper morphologies, grasp types, and physical realism.

Traditional grasp synthesis methods for collecting these datasets rely either on sampling-based algorithms~\citep{Miller2004,sampling_grasping_icra} or analytical techniques that evaluate force closure by solving (conic) optimization problems~\citep{liu2020new, grasp_analysis_as_lmi,tro_qualittative_test_grasp}.
While effective at a basic level, these approaches suffer from poor sample efficiency, especially when accurate physical modeling is required, and tend to converge to simple power grasps rather than diverse, task-appropriate configurations.
More recent gradient-based methods~\citep{liu2021synthesizing, turpin2022grasp, wang2023dexgraspnet} improve efficiency by optimizing differentiable approximations of grasp quality metrics, but often oversimplify critical physical properties, thereby limiting grasp realism and diversity.

To overcome these limitations, we propose a new framework for synthesizing large-scale, diverse, and physically grounded dexterous grasps.
First, we develop a differentiable force closure energy term that rigorously models grasp stability by considering frictional forces within an implicit Quadratic Program (QP) formulation.
In contrast to prior approaches~\citep{wang2023dexgraspnet,li2022gendexgrasp,turpin2023fastgraspd,liu2021synthesizing} that rely on simplified or relaxed metrics, our formulation preserves the core physical properties of grasp stability while remaining fully differentiable.
This enables gradient-based optimization without compromising physical accuracy and allows the synthesis of grasps that include refined manipulation strategies such as pinch and tri-finger precision grasps.
Second, we propose MALA*, an enhanced optimization and sampling strategy inspired by the Metropolis-Adjusted Langevin Algorithm (MALA).
MALA* dynamically rejects gradient steps that would lead to poor sample diversity, using information about the global distribution of existing samples.
This adjustment mitigates mode collapse during optimization and encourages exploration of a broader range of feasible and physically distinct grasps.
We extensively evaluate our approach and demonstrate significant improvements over existing techniques in terms of grasp diversity and stability.
Additionally, we release modular grasping environments for evaluation and reinforcement learning, built on IsaacLab~\citep{mittal2023orbit}.

Together, these advancements enable the efficient synthesis of large-scale, high-diversity grasp datasets.
To validate our approach, we perform detailed evaluations and generate a new dataset comprising grasps for 5,700 objects from DexGraspNet~\citep{wang2023dexgraspnet}.
The dataset includes annotations for five different grippers and spans three distinct grasp types, substantially expanding the diversity of available data for dexterous manipulation.

In summary, our contributions are: 
\begin{enumerate}[(i)]
    \item We improve upon existing analytical grasp generation methods using a more analytically rigorous force closure formulation.
    \item We introduce an adjusted optimization method, MALA*, that improves convergence and diversity.
    \item We generate large-scale grasp predictions across various objects, grippers, and grasp taxonomies.
\end{enumerate}

\section{Related Work}
\subsection{Analytical Grasp Synthesis}
Grasp synthesis aims to generate a diverse set of stable grasp predictions for known hand and object pairs. Traditional approaches to grasp synthesis primarily rely on sampling-based methods or sequences of linear programs~\citep{Miller2004, casas2024multigrippergraspdatasetroboticgrasping, Dai2018,wu2022learning}, which explore the space of possible hand poses to optimize a given grasp quality metric. However, these methods can be computationally expensive, particularly for high-degree-of-freedom (DoF) dexterous hands.  To improve efficiency, recent methods have shifted toward gradient-based optimization~\citep{liu2021synthesizing, li2022gendexgrasp, turpin2022grasp, turpin2023fastgraspd,chen2024springgrasp}. These approaches leverage differentiable grasp quality metrics, enabling faster convergence compared to sampling-based methods. However, this reliance on differentiability introduces limitations. Classical grasp quality metrics such as \( Q_1/ \varepsilon\)~\citep{Ferrari1992PlanningOG,Kirkpatrick1992,kappler_big_data_grasping}, which involve solving min-max optimization problems, are not inherently differentiable. To address this, recent works have proposed relaxed, differentiable formulations~\citep{liu2020deep, liu2021synthesizing}, \changed{bilevel optimization~\citep{li2023frogger,wu2022learning,chen2024bodex}} and sampling-based wrench space analysis~\citep{chen2024task}, which allow for gradient-based optimization. Despite their computational advantages, these relaxed formulations often introduce implicit assumptions that can oversimplify grasp stability. Notably, some methods fully neglect frictional forces~\citep{liu2021synthesizing, wang2023dexgraspnet}, which are critical for certain grasp configurations. \changed{
BilevelOpt~\citep{wu2022learning} incorporates the force closure condition into the constraint formulation of an optimization problem and relies on suboptimal initial solutions from IK solvers and contact points.
FRoGGeR~\citep{li2023frogger} employs a differentiable force closure metric similar to ours but constrains the interaction wrench coefficients to sum to one to prevent vanishing gradients.
Similarly, concurrent work such as BoDEX~\citep{chen2024bodex} enforces a bound on contact wrenches by requiring the sum of normal forces to exceed a predefined hyperparameter $\gamma$.}
Alternative approaches mitigate this by incorporating differentiable physics simulations, using velocity-based metrics to evaluate grasp quality~\citep{turpin2022grasp, turpin2023fastgraspd}, but they depend on simplified contact assumptions or relaxations, such as penalty-based contact modeling.
\subsection{Dexterous Grasping Datasets}
A wide range of datasets~\citep{casas2024multigrippergraspdatasetroboticgrasping, turpin2023fastgraspd, wang2023dexgraspnet, li2022gendexgrasp, zhang2025graspxl, 9560844, mahler2017dex, kappler_big_data_grasping,multi-fingan,columbia_database} provide synthesized grasp poses for rigid objects across different grippers.
These datasets vary in scale, from sets with a few hundred thousand grasps~\citep{liu2020deep, li2022gendexgrasp} to large-scale datasets containing a million or more grasp annotations~\citep{9560844, li2022gendexgrasp, wang2023dexgraspnet, casas2024multigrippergraspdatasetroboticgrasping}.
In terms of gripper diversity, many focus on parallel jaw grippers~\citep{9560844,shao2020unigrasp}, while few incorporate dexterous hands such as the Shadow and Allegro Hand~\citep{casas2024multigrippergraspdatasetroboticgrasping, turpin2023fastgraspd}.
Our dataset extends the diversity of grippers by providing annotations for the Psyonic Ability Hand, a commonly used research hand.
We additionally focus on increasing the diversity of generated grasps per object, both in joint entropy and in taxonomy.

\section{Preliminaries}

\oursubsubsection{Positive Spanning Sets}
As formally defined in literature~\citep{posSpans}, positive spanning sets are characterized as follows:
\begin{definition}[Positive span]
The positive span of a finite set of vectors $\mathcal{S} = \{v_1, ..., v_k\}$, denoted $\PSpan{\mathcal{S}}$, is given by:$
\ \ \PSpan{\mathcal{S}}  := \{ \lambda_1v_1 + ... + \lambda_kv_k \ : \ \lambda_i \geq 0\ \forall i = 1,...,k\}.
$
% Similarly, the positive span of a matrix $W \in \mathbb{R}^{k \times c}$ denoted by  $\PSpan{W}$, is given by the positive span of the set of all of its column vectors.
\end{definition}
Additionally, \citep{posSpans} introduce the following theorem about the range of positive spanning sets:
\begin{theorem}[Range of positive spanning sets]
\label{thm:range}
Suppose $S = \{v_1,...,v_k \} \subseteq \mathbb{R}^n$, with $v_i \neq 0$, \underline{linearly} spans the subspace V of $\mathbb{{R}}^n$. The following statements are equivalent: \\
\hspace*{4mm}(i) The set S \underline{positively} spans V \\
\hspace*{4mm}(ii)  $\exists \ \alpha_1,...,\alpha_k > 0$ such that $\sum_{i=1}^k \alpha_i v_i= 0$ \\
\hspace*{4mm}(iii) $\exists \ \gamma_1,...,\gamma_k \geq 0$ such that $\sum_{i=1}^k \gamma_i v_i= - \sum_{i=1}^{k}v_i$ 
\end{theorem}
Note that Theorem~\ref{thm:range}~-~(ii) is well known and often referred to as ``\emph{the origin must lie in the convex hull of the spanning set}'' in the literature.

\begin{figure*}
    \centering
    \vspace{-1cm}
    \includegraphics[trim={0 3.3cm 0 0},clip,width=1.0\linewidth]{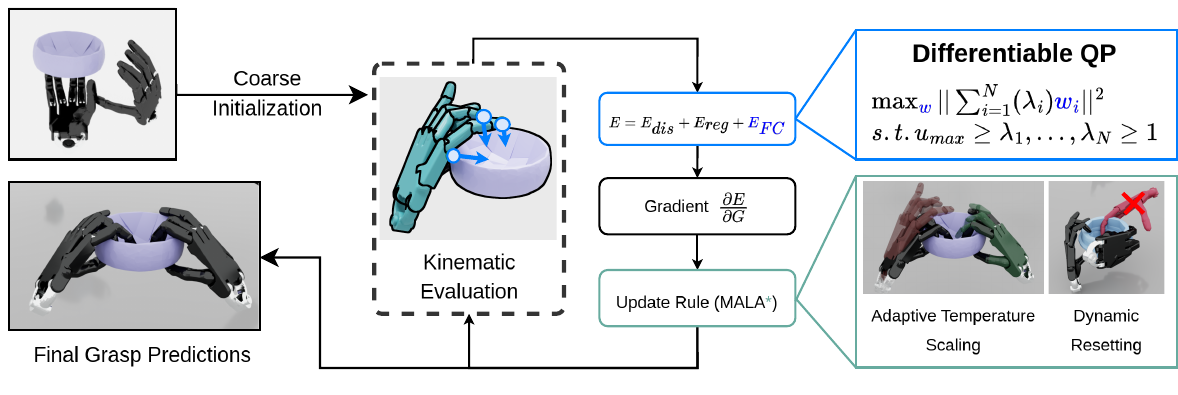}
    \caption{\textbf{Overview of our Grasp Generation Method:} Starting from a coarse initialization, grasp candidates are optimized by minimizing a composite energy function composed of distance, regularization, and force closure terms.
    The force closure energy is computed via a differentiable quadratic program (QP) that ensures the contact wrenches positively span $\mathbb{R}^6$.
    Gradients of the energy are used to iteratively refine grasp poses through our modified Metropolis-adjusted Langevin (MALA$^*$) algorithm, which dynamically adapts updates and resets based on the energy landscape.
    Grasp proposals that significantly underperform relative to the current energy distribution are reset (\emph{Dynamic Resetting}), and the acceptance probability of each gradient step is modulated according to the proposal's relative energy performance (\emph{Adaptive Temperature Scaling}). After convergence, the method produces diverse and physically stable grasp predictions.
    }
    \label{fig:method_overview}
\end{figure*}

\oursubsubsection{Form and Force Closure}
Given a set of contact points \( \{x_i\}_{i=1}^N \), where \( x_i \in \mathbb{R}^3 \), a grasp is defined to be in \emph{form closure} if the static contact points, together with their associated frictionless interaction forces \( \{f_i\}_{i=1}^N \), \( f_i \in \mathbb{R}^3 \), completely constrain all possible motions of the object.
More formally, this condition is satisfied if the set of contact wrenches $\mathcal{W} = \{w_i\}_{i=1}^N$ with $w_i = [f_i; (f_i \times x_i)] \in \mathbb{R}^{6\times 1}$, constructed from the contact points $\{x_i\}_{i=1}^N$ and the corresponding interaction forces $\{f_i\}_{i=1}^N$, positively spans the entire wrench space \( \mathbb{R}^6 \): $\PSpan{\mathcal{W}} = \mathbb{R}^6.$
% \begin{equation}
% \end{equation}

Furthermore, \emph{force closure} requires that the grasp can resist any arbitrary external wrench applied to the object by additionally leveraging frictional properties at contact points, i.e. the set of wrenches generated by the contact forces $\mathcal{W_{FC}} = \{ \hat w | \hat w \in \bigcup_{i=1}^N WC\left(w_i\right)  \}$ positively span the entire wrench space \( \mathbb{R}^6 \):
% Unlike \emph{form closure}, which relies purely on the geometric arrangement of contact points, force closure considers the interaction wrenches $w$ constrained within the wrench cones $WC(w)$. In other words, a grasp is in force closure if the set of wrenches generated by the contact forces $\mathcal{W_{FC}} = \{ \hat w | \hat w \in \bigcup_{i=1}^N WC\left(w_i\right)  \}$ positively span the entire wrench space \( \mathbb{R}^6 \),
% \begin{equation}
\refstepcounter{equation} \(
\label{eq:force-closure}
\PSpan{\mathcal{W_{FC}}} = \mathbb{R}^6
\)\hfill ~(\theequation).

% \end{equation}
\changed{Finally, we define the \emph{Wrench Matrix} $W_{FC} = [\, \hat w_1, \, \hat w_2, \, \dots, \, \hat w_N\,] \in \mathbb{R}^{ 6 \times N}$ as the matrix whose $j$-th column is the contact wrench $\hat w_j \in \mathcal{W_{FC}}$, where $N$ is the total number of interaction wrenches. Force closure is satisfied iff the positive span of the columns of $W_{FC}$ spans $\mathbb{R}^6$.
}
% Note that force closure is a stronger assumption than form closure. Therefore, any grasp in form closure is also in force closure. 

% The form of $\left\lfloor x_i\right\rfloor_{\times}$ensures the cross product $\left\lfloor x_i\right\rfloor_{\times} f_i=x_i \times$

% $$
% \begin{aligned}
% G G^{\prime} & \geq \epsilon I_{6 \times 6}, \\
% G f & =0 \\
% f_i^T c_i & >\frac{1}{\sqrt{\mu^2+1}}\left|f_i\right|, \\
% x_i & \in S
% \end{aligned}
% $$
% where $S$ is the object surface, and
% $$
% \begin{aligned}
% G & =\left[\begin{array}{cccc}
% I_{3 \times 3} & I_{3 \times 3} & \ldots & I_{3 \times 3} \\
% \left\lfloor x_1\right\rfloor_{\times} & \left\lfloor x_2\right\rfloor_{\times} & \ldots & \left\lfloor x_n\right\rfloor_{\times}
% \end{array}\right], \\
% \left\lfloor x_i\right\rfloor_{\times} & =\left[\begin{array}{ccc}
% 0 & -x_i^{(3)} & x_i^{(2)} \\
% x_i^{(3)} & 0 & -x_i^{(1)} \\
% -x_i^{(2)} & x_i^{(1)} & 0
% \end{array}\right] .
% \end{aligned}
% $$

\oursubsubsection{Grasp Representation}
For a given robotic gripper with $n_q$ joints, we parameterize each grasp as $G = ({\chi}, {q}, {\Delta q} )$, where ${\chi} \in SE(3)$ denotes the wrist pose, ${q} \in \mathbb{R}^{n_q}$ the joint positions of the $n_q$ joints of the robot and ${\tau_q} \in \mathbb{R}^{n_q}$ the desired joint torques.
Furthermore, each grasp is associated with a set of $N_C$ contact points $\mathcal{C} = \{ c_i\}_{i=1}^{N_c}$, given in the object frame, and their corresponding object normals $\mathcal{N} = \{ n_i\}_{i=1}^{N_c}$. 
Given these contact points, we calculate the set of interaction wrenches as $\mathcal{W} = \{ w_i\}_{i=1}^{n_c}$, where $w_i = [n_i, c_i \times n_i] \in \mathbb{R}^6$. 
Finally, when considering friction, we rely on the four-sided pyramid under-approximation of the friction cone $FC(f_i, c_i, n_i)$~\citep{modern_robotics}.  %, which is defined as the set of cone edges in the local contact frame $(\mu , 0, f_i),
% (-\mu, 0, 1), (0, \mu, 1)$, and $(0, -\mu, 1)$ where $\mu$ denotes the static coulomb friction coefficient.

% $\mathcal{W_{FC}}$ then denotes the set of wrenches generated by the contact forces when considering the friction cones, i.e., $\mathcal{W_{FC}} = \bigcup_i \{ w_i = [\hat{f}_i]\in FC(f_i, c_i, n_i)\}$ 
% \newpage
\section{Method}

% \subsection{Problem Formulation and Notation}
Given an accurate 3D model of an object and a dexterous hand, we aim to synthesize a diverse set of feasible grasp configurations $\mathcal{G} = \{ G_i\}_{i=1}^N$ that satisfy the force closure property in Eq.~\eqref{eq:force-closure}.
To accomplish this, we first generate a diverse initial set of feasible grasps and then refine them through an analytical optimization process that maximizes the span of the associated friction cones.
An overview of our method is depicted in Fig.~\ref{fig:method_overview}, and final grasp predictions are shown in Fig.~\ref{fig:title-fig}.

\subsection{Analytical Grasp Generation}
\label{subsec:analytical_optimization}
We build upon a grasp generation pipeline commonly employed in gradient-based grasp synthesis~\citep{turpin2023fastgraspd, turpin2022grasp, liu2021synthesizing, wang2023dexgraspnet}, following an approach similar to DexGraspNet~\citep{wang2023dexgraspnet}, which we briefly review below.
Each gripper is associated with a predefined set of potential contact points $\mathcal{C}$, of which only a subset $\mathcal{C}'$ is active at a time (i.e., considered to be in contact with the object).
Given a grasp-object configuration, the energy of the grasp is computed as
% In contrast to our approach, DexGraspNet does not consider the desired  joint actions directly. % and therefore their grasp definition can be formulated as $G'=({\chi},{q}, {x'}$). 
% They then calculate the energy of the resulting grasp as:
\begin{equation}
E = E_{FC} + w_{dis}E_{dis} + w_{reg}E_{reg},
\end{equation}
with $E_{FC}$ being a force closure metric, $E_{dis}$ denotes the distance of the active contact points ($\mathcal{C}'$) to the object surface, and $E_{reg}$ consists of regularization energies ($E_{pen}, E_{joints}, E_{spen}$) which are used to prevent the grasp from penetrating the object, itself, or exceeding the joint limits, respectively.

\oursubsubsection{Force Closure Metric} 
DexGraspNet~\citep{wang2023dexgraspnet} assumes the grasp to be frictionless and that the force applied at each contact point has equal magnitude, which does not accurately model real-world hardware, where fingers can be actuated separately. Formally, under these assumptions, their force closure metric becomes a form closure assumption and a simplification of Theorem~\ref{thm:range}~-~(ii) where $\alpha_i = 1$. This results in $E_{FC} = ||\sum_{i \leq |\mathcal{C}'|} w_i||_2 \rightarrow  0$,
i.e., the sum of the column vectors of the wrench space must sum up to zero. %Furthermore, DexGraspNet relies on a simplified gradient descent algorithm of the Metropolis-adjusted Langevin algorithm (MALA), introduced in~\citep{liu2021synthesizing} and depicted in~\ref{alg:mala_optim}.

A more general formulation of a force closure metric could be achieved by directly using Theorem~\ref{thm:range} - (ii) to find the optimal coefficients $\alpha_i$ for the current contact points $\mathcal{C}'$, i.e., 
\begin{align}
    E_{FC} &= ||\sum_{i \leq |\mathcal{C}'|} \alpha_i w_i||_2   \ \  \text{s.t.}\ \alpha_i > 0\ \forall i = 1,...,n_c.
\label{eq:degenerate-fc}
\end{align}
While this approach would be more general, it is challenging to include as an optimization term, since the error term $E_{FC}$ can become arbitrarily small for arbitrarily small choices of $\alpha_i$, leading to vanishing gradients. Hence, we propose relying on the more robust formulation in Theorem \ref{thm:range} - (iii) to ensure that the grasp is in force closure:
\begin{align}
    E_{FC} &= ||\sum_{i \leq |\mathcal{C}'|} \gamma_i w_i  + \sum_{i \leq |\mathcal{C}'|} w_i||_2 = ||\sum_{i \leq |\mathcal{C}'|} \hat\gamma_i w_i||_2 \ \  \text{s.t.}\ u \geq \hat\gamma_i \geq 1\ \forall i = 1,...,n_c,
\end{align}
where $u$ is an upper bound on the interaction force for each contact point.
While this is not strictly necessary for force closure, it ensures that the required interaction forces remain bounded, which is necessary when using a real robot with torque limits.
The resulting coefficients $\hat\gamma_i$ directly reflect the force applied at each contact point to keep the object stationary, as opposed to the more degenerate formulation in Eq.~\eqref{eq:degenerate-fc}, which incentivizes the fingers to apply a minimal amount of force.
Note that this formulation is equivalent to the force closure metric from \citet{liu2021synthesizing} if $\hat\gamma_i = 1$ is chosen for all $i$.

%Furthermore, it poses a better suited optimization problem, as the error term $E_{FC}$ cannot become arbitrarily small due to the constraint $\hat\gamma_i \geq 1$.

Furthermore, we note that the force closure metric is implicitly defined through an optimization problem and has imposed hard constraints (introduced by the bounds on $\hat\gamma$). This makes propagating the gradient of the entire problem more challenging.
We propose and evaluate two different approaches to make the optimization problem differentiable:

\begin{enumerate}
    \item \emph{Unconstrained Optimization}: 
    \label{sec:nonlinear_prob}
    By introducing barrier functions, the energy term can be converted into an unconstrained optimization problem. 
    Hence, the energy term $E_{FC}$ is replaced with $E_{FC} + \texttt{barrier}(\hat\gamma)$ and the resulting unconstrained optimization problem is solved using a trust region-based optimizer (Powell's Hybrid Method)~\citep{Powell1970}. 
    \item \emph{Differential QP}: We formulate the optimization problem as a Quadratic Program (QP):
    \begin{alignat*}{2}
        E_{FC} &= \min_{z} \frac{1}{2} z^T H z + g^T z \ \ \ \text{s.t.}  \ A z \geq b,  \\
        H &= W_{FC}^TW_{FC},\ g = 0, \ b = [1_{n_c}; u_{n_c}],\\
        z &= [\hat\gamma_1,..., \hat\gamma_{N_c}],\  A= \texttt{diag}(1_{n_c\times n_c},-1_{n_c\times n_c}).
    \end{alignat*}
    The gradients with respect to the input parameters $z$ can be calculated by differentiating the Karush–Kuhn–Tucker (KKT) conditions of the QP, as shown by Amos \textit{et al.}~\citep{qpth}.
    % We use this method to backpropagate the gradients through the QP.
\end{enumerate}

Finally, Theorem~\ref{thm:range} guarantees force closure only if the wrench space linearly spans $\mathbb{{R}}^6$.
The original metric by Liu \textit{et al.}~\citep{liu2021synthesizing} ensures this by additionally optimizing over the smallest singular value of the wrench matrix $W_{FC}$.
However, subsequent works~\citep{wang2023dexgraspnet, li2022gendexgrasp} do not include this in their energy formulation and hence lose force-closure guarantees.
To ensure the wrench matrix forms a linear basis, we propose to scale the energy term by 
% $e^{-\texttt{det}(\mathcal{W_{FC}})}$ where $\texttt{det}(\mathcal{W_{FC})}$ 
\changed{$e^{-\prod_i\sigma_i(W_{FC})}$, where $\sigma_i$ are the singular values of the wrench matrix. All $\sigma_i > 0$ ensures full rank, satisfying Theorem~\ref{thm:range}, while maximizing their product maximizes the wrench space volume.}

Therefore, we define the final energy term as:
\begin{equation}\vspace{2mm}
    E_{FC} = ||\sum_{i \leq |\mathcal{C}'|} \hat\gamma_i w_i||_2 \cdot \changed{e^{-\prod_i\sigma_i(W_{FC})}}      \ \ \ \text{s.t.}  \ \hat\gamma_i \geq 1 \ \forall i = 1,...,n_c.
\end{equation}\vspace{-5mm}
\oursubsubsection{Distribution-based MALA Optimizer}
We often observe that certain grasp proposals get stuck in local minima (i.e., bad grasp configurations) during optimization. Escaping these minima proves challenging since the optimization is rolled out for each grasp independently without any knowledge of the current loss landscape. However, such knowledge (and therefore the detection of local minima) can be made available by incorporating additional information from the full grasp distribution used for optimization at the current time step $t$.
We therefore introduce the following changes to the optimization subroutine\footnote{The full algorithm is described in the Appendix - Alg.\ref{alg:mala_optim}}, which is applied to the grasp distribution at each iteration $k$:
$\mathcal{G}^{(k)} = \{ G^{(k)}_{0}, ..., G^{(k)}_{N}\}$ with energies $E^{(t)} = \{E_0^{(k)}, ...., E_N^{(k)}\}$:
% The final algorithm (with our changes highlighted in \textcolor{blue}{blue}) is depicted in Alg.~\ref{alg:mala_optim}.
% \begin{enumerate}
    % \item

    \textbf{{Dynamic Resetting}}:
    To prevent grasp starvation (i.e., grasp poses converging to local minima), we reset the optimizer for grasps that are performing significantly worse compared to the overall grasp distribution. After every $n_{reset}$ steps, we fit a normal distribution, parameterized by $\mathcal{N}_E(\mu, \sigma)$, over the current set of energies $E^{(k)}$. We then re-initialize all grasp poses with $\Phi_E(E_i; \mu, \sigma) \leq p_{th}$, i.e., the grasps that are within the lowest quantile of the fitted normal distribution, which is calculated based on the cumulative distribution function $\Phi_E$ and the upper probability threshold $p_{th}$ are re-initialized.
    
    \textbf{Adaptive Temperature Scaling}:
    We update the Metropolis--Hastings acceptance criterion for accepting gradient steps.
    In the original formulation, a gradient step that introduces an energy change $\Delta E$ is accepted with probability $p \sim e^{-\Delta E/T}$.
    We propose conditioning the temperature $T$ on the energy distribution, i.e., if a grasp performs significantly worse than the overall distribution, we increase the temperature for that grasp.
    Hence, we change the acceptance criterion to $p \sim e^{-\Delta E / T_i}$, where $T_i$ is the temperature for grasp $i$, and calculated using the following formula:
    $T_i =  T\cdot (1  + \Phi_E(E_i))$.    
    % \item \emph{Adam Optimizer}:
    % We adapt the current optimizer to use the Adam optimizer \citep{adam2017}, which allows for more efficient optimization and better convergence properties.
We refer the reader to Appendix~\ref{alg:mala_optim} for the complete formulation and pseudocode of the optimizer.
%\end{enumerate}
\section{Results}
\begin{figure}
    \centering
    % \centering
    \vspace{-1cm}
    \includegraphics[width=0.7\linewidth]{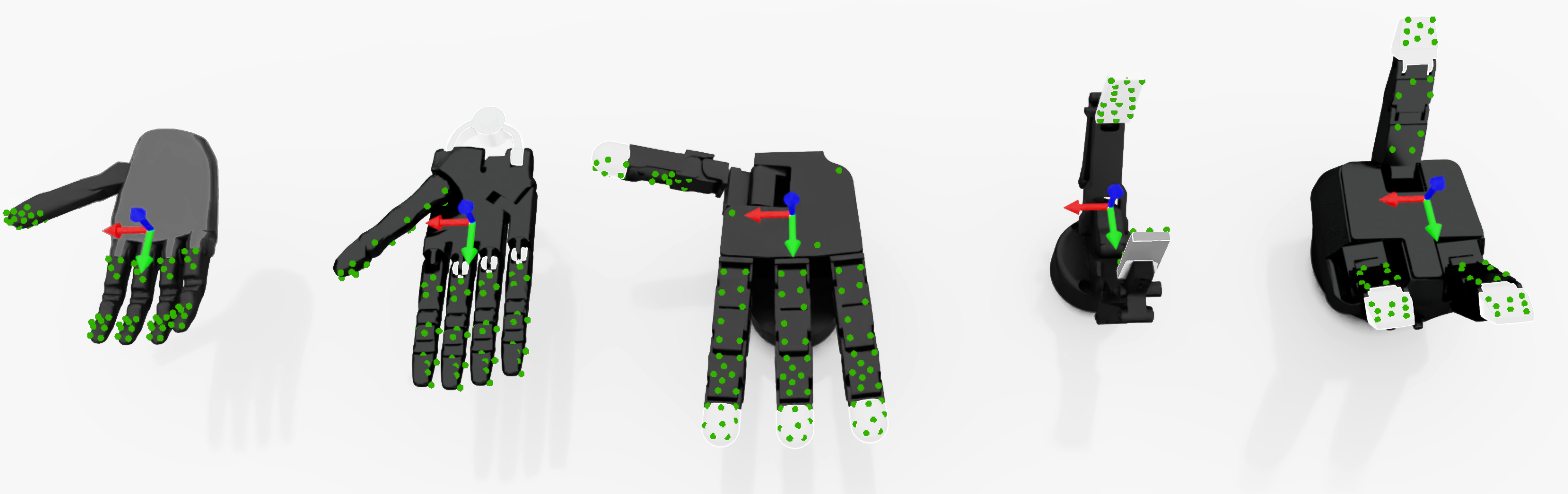}
    \caption{\textbf{Overview of the Five Robotic Grippers used for Evaluation:} From left to right, Psyonic Ability Hand, Shadow Hand, Allegro Hand, Robotiq2f140 and Robotiq3F. We show the standardized wrist frame convention with axes indicating thumb (x-axis), fingers (y-axis), and palm normal (z-axis) directions. Randomly sampled contact points from the manually defined contact meshes are shown in green.}    
    \label{fig:hands_overview}
    \vspace{-3mm}
    
\end{figure}

\oursubsubsection{Main Evaluation Metrics}
To assess the performance of our method, we introduce the following evaluation metrics:\\
% \begin{itemize}\vspace{-3mm}
\emph{Unique Grasp Rate (UGR)} captures the number of distinct, stable grasps that the method can generate. This metric is calculated by discretizing the successful grasp poses using three different resolutions $\delta_r=2\,\text{cm}$, $\delta_\phi=4^\circ$, and $\delta_q=1.15^\circ$ for the position, orientation (Euler angles), and joint states, respectively.
We then calculate the number of unique grasps and report the UGR as the ratio of unique successful grasps over the total number of grasps for each object.\\
\emph{Entropy (H)}, a common measure for the diversity of grasp configurations is the mean entropy of the finger joints. We extend this metric to also include the position and orientation of the grasp configuration since, depending on the task at hand, different grasp positions and orientations might be more suitable. Note that calculating the entropy of an SE(3) rotation is non-trivial. We refer the reader to Appendix~\ref{app:entropy} for the full derivation.
% \end{itemize}

\oursubsubsection{Evaluation}
We assess the stability of each grasp proposal using Isaac Lab~\citep{mittal2023orbit} and an evaluation method as introduced in \citep{wang2023dexgraspnet}.
Specifically, while grasping, forces are applied to the objects in the six canonical world directions.
We assess stability by checking that the center of mass (CoM) of the objects stays within a sphere of radius $3\,\text{cm}$ for at least one of the principal axis pairs ($\pm x$, $\pm y$, $\pm z$). 

\oursubsubsection{Dataset}
We convert the dataset introduced by \citet{wang2023dexgraspnet}, comprising 5,700 objects, into the Universal Scene Description (USD) format to enable direct use within Isaac Sim.
To improve collision handling, we recompute the collision meshes using Quad Remesher~\citep{exoside}, and set the refined meshes as the new collision geometry for GPU-accelerated signed distance field (SDF) collision checking.
For evaluation purposes, we randomly select a subset of 50 objects from the dataset to form our test set.
All reported results are based on this test set.

\begin{table*}[t]\vspace{-5mm}
    \caption{\textbf{\textsc{Synthesized Grasp Poses}}: 
    We compare the influence of different analytical optimization schemes on the final grasp quality and diversity metrics, for 4 and 12 contact points respectively, using our test set of 50 objects.
	We report the unique grasp rate (\emph{UGR}) as the percentage of successful unique grasps over all generated unique grasps, the joint entropy (H) of the successful grasp distribution, and the average max penetration depth ($D$).
	All evaluations are performed in a no-gravity environment with $5\,\text{N}$ force.
    % \todo{Check DexGraspNet}
    }
\label{tab:main_results}
\setlength{\tabcolsep}{3pt}
\centering
        
\resizebox{\textwidth}{!}{
	\begin{tabular}{l|l|l|l|cc|cc|cc|cc|cc||cc|c}
		\toprule
		        
		\ \ \ \  & Grasp  & \multirow{ 2}{*}{Method}  &  \multirow{ 2}{*}{Optimizer} &  \multicolumn{2}{c|}{Allegro} &  \multicolumn{2}{c|}{Shadow Hand}&  \multicolumn{2}{c|}{Robotiq3f} &  \multicolumn{2}{c|}{Robotiq2f} &  \multicolumn{2}{c|}{Ability Hand}  & \multicolumn{3}{c}{Overall} \\
		    
		& Type  &   & &UGR $\uparrow$ 
		%& SR$\uparrow$ 
		&   H  $\uparrow$ 
		&   UGR $\uparrow$ 
		% & SR$\uparrow$ 
		&   H $\uparrow$ 
		& UGR $\uparrow$ 
		% & SR $\uparrow$ 
		&   H $\uparrow$ 
		&  UGR $\uparrow$ 
		% & SR$\uparrow$ 
		&   H $\uparrow$ 
		&  UGR $\uparrow$ 
		% & SR$\uparrow$ 
		&   H$\uparrow$ 
		&  UGR $\uparrow$ 
		% & SR$\uparrow$ 
		&   H $\uparrow$ & D $\downarrow$\tiny{[mm]}\\
		\midrule

		\parbox[t]{0mm}{\multirow{3}{*}{\rotatebox[origin=c]{90}{4 Ctc.}}}  &

		\parbox[t]{0mm}{\multirow{3}{*}{
		
		\includegraphics[width=8mm]{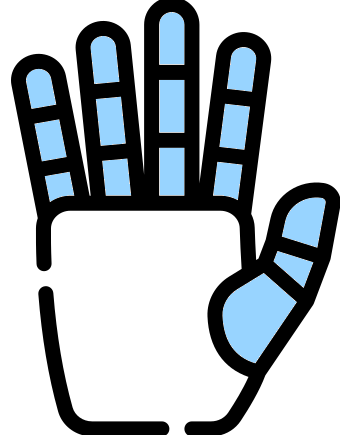} \vspace{2mm}
		% \rotatebox[origin=c]{90}{Power}
		}
		}   
		% POWER UGR 4 CONTACTS
		& DexGraspNet~\citep{wang2023dexgraspnet} & MALA & 67\% & \underline{3.5} & \underline{38\%} & \underline{2.9} & 26\% & 2.3 & 28\% & 3 & \underline{15\%} & \underline{1.6} & 35\% & \underline{2.7} & 1.0\\

% & &  DexGraspNet & MALA & 67\% & 3.5 & 38\% & 2.9 & 26\% & 2.3 & 28\% & 3 & 15\% & 1.6 & 35\% & 2.7 \\
        
% & & DexGraspNet & MALA$^*$ & 71\% & 3.5 & 36\% & 2.9 & 34\% & 2.5 & \underline{32\%} & 3.1 & 14\% & 1.5 & 38\% & 2.7 \\

		  &   & DexGraspNet~\citep{wang2023dexgraspnet} & MALA$^*$ & \underline{71\%} & \underline{3.5} & 36\%             & \underline{2.9}             & \underline{34\%} & \underline{2.5} & \underline{32\%}    & \underline{3.1} & 14\%                     & 1.5             & \underline{38\%} & \underline{2.7} & \textbf{0.9} \\
		%  &   & \OurMethod (ours)                       & MALA$^*$ & \textbf{73\%}    & \textbf{3.6}    & \textbf{39\%}    & \textbf{3}      & \textbf{35\%}    & \textbf{2.6}    & \textbf{32\%}    & \textbf{3.2}    & \textbf{20\%}            & \textbf{2}      & \textbf{40\%}    & \textbf{2.9}    \\
		&& \OurMethod (ours)     & MALA$^*$ & \textbf{74\%} & \textbf{3.6} & \textbf{49\%} & \textbf{3.2} & \textbf{49\%} & \textbf{2.9} & \textbf{34\%} & \textbf{3.3} & \underline{19\%} & \textbf{1.8} & \textbf{45\%} & \textbf{3.0} & \textbf{0.9} \\

		\midrule
		\midrule

		\parbox[t]{0mm}{\multirow{14}{*}{\rotatebox[origin=c]{90}{12 Contacts}}}  
		 
		& \parbox[t]{0mm}{\multirow{3}{*}{
		\includegraphics[width=8mm]{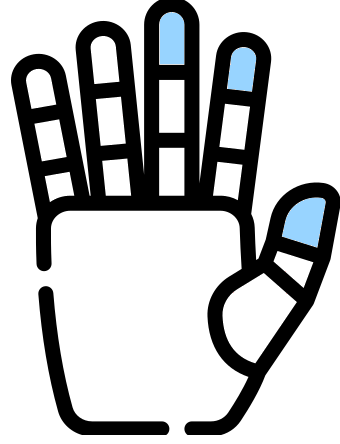} \vspace{2mm}
		% \rotatebox[origin=c]{90}{Prec.}
		}}   
		
		& DexGraspNet~\citep{wang2023dexgraspnet} & MALA & 37\% & 2.7 & 14\% & 1.6 & 5\% & 0.6 & -- & -- & 14\% & 1.5 & 18\% & 1.6 & \textbf{1.9} \\
		  &   & GenDexGrasp~\citep{li2022gendexgrasp}   & MALA$^*$ & 48\%             & 3.1             & \underline{23\%} & 2.2             & \textbf{9\%}     & \textbf{1.1} & --               & --              & \underline{18\%}         & \underline{1.8} & \underline{25\%} & \underline{2.0} & 2.6 \\
		  &   & TDG~\citep{chen2024task}                & MALA$^*$ & \underline{49\%} & \underline{3.3} & \underline{23\%} & \underline{2.3} & 8\%  & 0.9            & --               & --              & 16\%                     & 1.6             & 24\%             & \underline{2.0}   &  2.7           \\
		  &   & \OurMethod (ours)                       & MALA$^*$ & \textbf{55\%}    & \textbf{3.3}    & \textbf{29\%}    & \textbf{2.5}    & \textbf{9\%}     & \textbf{1.1}    & --               & --              & \textbf{26\%}            & \textbf{2.3}    & \textbf{30\%}    & \textbf{2.3} & \underline{2.3} \\
		
		\cmidrule{2-17}
		 
		&
		\parbox[t]{0mm}{\multirow{3}{*}{
		% \rotatebox[origin=c]{90}{Pinch}
		   
		\includegraphics[width=8mm]{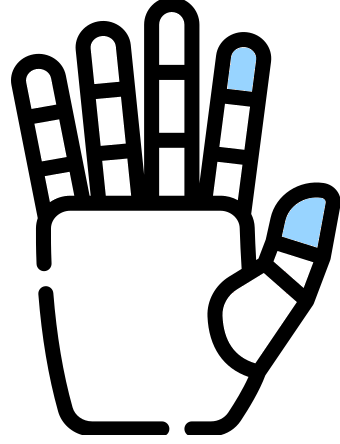} 
		}}   
		
		% PINCH
		& DexGraspNet~\citep{wang2023dexgraspnet}  & MALA & 31\% & 2.5 & 10\% & 1.3 & \underline{4\%} & \underline{0.5} & -- & -- & 11\% & 1.2 & 14\% & 1.4 & \textbf{1.8} \\
		  &   & GenDexGrasp~\citep{li2022gendexgrasp}   & MALA$^*$ & \underline{44\%} & 2.9             & \underline{14\%} & \underline{1.6} & \underline{4\%}  & 0.4            & --               & --              & \underline{17\%}         & \underline{1.8} & \underline{20\%} & \underline{1.7}     & 2.6        \\
		
		  &   & TDG~\citep{chen2024task}                & MALA$^*$ & 42\%             & \underline{3.0}   & 13\%             & 1.4             & \textbf{5\%}     & \textbf{0.6}   & --               & --              & 16\%                     &1.7           & 19\%             & \underline{1.7} & 2.7 \\
		
		  &   & \OurMethod (ours)                       & MALA$^*$ & \textbf{49\%}    & \textbf{3.1}    & \textbf{17\%}    & \textbf{1.8}    & \underline{4\%}  & \underline{0.5}            & --               & --              & \textbf{23\%}            & \textbf{2.1}    & \textbf{23\%}    & \textbf{1.9}   & \underline{2.3} \\
		
		\cmidrule{2-17}
		&
		\parbox[t]{0mm}{\multirow{3}{*}{
		   
		\includegraphics[width=8mm]{imgs/icons/robotic-hand_5fingers.png} \vspace{2mm}
		% \rotatebox[origin=c]{90}{Power}}
		}   }
		& MultiGripper~\citep{casas2024multigrippergraspdatasetroboticgrasping} & GraspIt! \citep{Miller2004}  & 62\% & 3.3 &  17\% & 1.6 & 35\% & 2.6 & - & - & - &- & -& - & --\\

		  &   & DexGraspNet~\citep{wang2023dexgraspnet} & MALA   & 66\%             & 3.5             & 35\%             & 2.9             & 33\%             & 2.6             & 30\% & 3.2             & 18\%                     & 1.8             & 37\%             & 2.8              & \underline{1.6}\\
          
 & & DexGraspNet~\citep{wang2023dexgraspnet} & MALA$^*$ & 68\% & 3.5 & 41\% & 3.0 & 40\% & 2.8 & \underline{31\%} & 3.2 & 18\% & 1.7 & 40\% & 2.9 &  \underline{1.6}\\
		  &   & GenDexGrasp~\citep{li2022gendexgrasp}   & MALA$^*$ & \underline{74\%} & \textbf{3.6} & \underline{47\%} & \underline{3.2}             & 51\%             & \underline{2.9} & \textbf{32\%}    & \textbf{3.4}    & 23\%                     & \underline{2.2} & \underline{45\%} & \underline{3.0} & 1.7  \\
		  &   & TDG~\citep{chen2024task}                & MALA$^*$ & 69\%             & 3.5             & \underline{47\%} & \underline{3.2} & \underline{54\%} & \underline{2.9} & 28\%             & \underline{3.3} & \underline{25\%}         & \underline{2.2}            & \underline{45\%} & \underline{3.0}      & 1.7       \\
		  &   & \OurMethod(ours)                        & MALA$^*$ & \textbf{76\%}    & \textbf{3.6}    & \textbf{63\%}    & \textbf{3.5}    & \textbf{57\%}    & \textbf{3.0}      & 29\%             & 3.2             & \textbf{35\%}            & \textbf{2.6}    & \textbf{52\%}    & \textbf{3.2}   & \textbf{1.4}   \\
		\bottomrule
	\end{tabular} }
    \vspace{-5mm}
\end{table*} 
\oursubsubsection{Main Results}
We report our main results in Table~\ref{tab:main_results}, comparing grasp quality across different contact points, gripper types, and baseline methods when exposed to a $5\,\text{N}$ disturbance force.
We report the Unique Successful Grasp Rate (UGR), entropy (H), and average maximum penetration depth evaluated per gripper. 
We compare our method against four baselines, namely DexGraspNet~\citep{wang2023dexgraspnet}, GenDexGrasp~\citep{li2022gendexgrasp}, TDG~\citep{chen2024task}, and MultiGripperDataset~\citep{casas2024multigrippergraspdatasetroboticgrasping}. Implementation details for all methods are provided in Appendix~\ref{app:impl_details}.
Across all settings, our method (\OurMethod) consistently outperforms prior work both in grasp entropy (H) and unique grasp rate (UGR). 
We evaluate our approach under two settings, restricting the number of active contact points $|\mathcal{C}'|$ to either 4 or 12, and observe that our method scales better with a larger number of contact points, albeit at the cost of slightly larger penetration depths.
This trend is expected, as increasing the number of contact points not only promotes greater contact area but also facilitates easier satisfaction of the force closure condition.
We further observe that the addition of our modified optimizer, MALA$^*$, improves the overall UGR as well as the entropy. This effect is more pronounced in the 12 contact setting and discussed in Tab.~\ref{tab:ablations}.

The Robotiq2f gripper denotes the only case where our method does not significantly outperform existing methods. We largely attribute this to the fact that the gripper only consists of two finger pads that are always parallel, and the influence of the grasp metric becomes less important. Additionally, the UGR for this gripper tends to collapse, as many grasps converge to similar gripper widths when interacting with object parts that have uniform geometries.
However, even though the unique grasp rate is limited, the raw grasp success rate (reported in Appendix~\ref{app:full_results}) remains high (above 80\%), demonstrating that the generated grasps are still physically plausible and stable despite reduced diversity.
Furthermore, when evaluating different grasp taxonomies, we observe a substantial decline in the number of unique successful grasps, particularly for the Robotiq 3F gripper. In this case, fewer than 10\% of the optimized grasps remain both successful and unique, corresponding to approximately three valid grasps per object.
In general, we primarily attribute this degradation in performance to the reduced stability of pinch and precision grasps, which inherently involve a limited number of contact points.
However, as shown in Appendix~\ref{app:full_results}, when the external disturbance force is limited to $1\,\text{N}$, the unique grasp rate improves significantly, reaching 41\% for pinch and 49\% for precision grasps.
For completeness, we report all metrics across all disturbance forces in Appendix~\ref{app:full_results} and additionally list the raw grasp success rates for all cases.

\begin{wrapfigure}{r}{0.55\textwidth}
\vspace{-8mm}
  \begin{center}
   \includegraphics[trim={3mm 12mm 3mm 0},clip,width=0.54\textwidth]{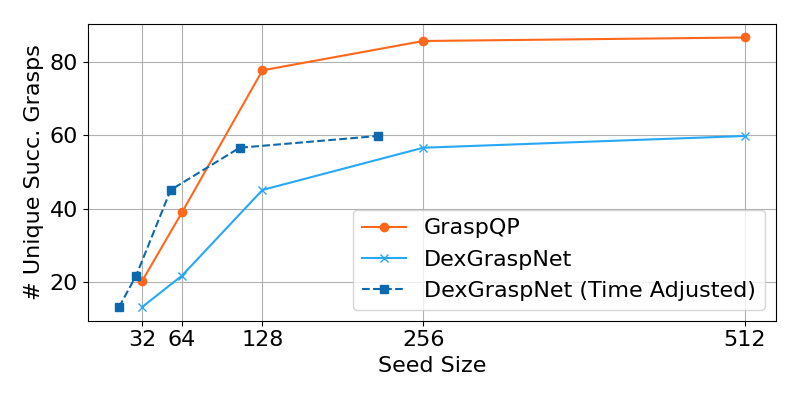}
  \end{center} 
  \caption{
  \changed{
\textbf{Unique Successful Grasps (UGR) vs. Seed Size}. Average UGR per object w.r.t. the number of initialized grasps (seeds). DexGraspNet yields more grasps under tight time constraints (Time Adjusted), but GraspQP scales more effectively, reaching ~80 unique grasps with only 128 seeds. DexGraspNet saturates at ~60 even with 512 seeds, indicating a diversity bottleneck that more samples cannot overcome.
  }
  } \vspace{-3mm}
  \label{fig:perf-over-seeds}
\end{wrapfigure}

\changed{Our approach is slower than the baseline (3.4 s vs. 1.15 s per grasp on the 24-DoF Shadow Hand) but is intended for offline dataset generation, where speed is less critical. The extra computation results in significantly more diverse grasps. As shown in Figure~\ref{fig:perf-over-seeds}, it may produce fewer grasps under tight time limits but ultimately reaches far more unique grasps (about 80 with 128 seeds) compared to the baseline’s saturation at ~60 even with 512 seeded grasps.}

\oursubsubsection{Ablation Studies}
We conduct ablation studies to assess the impact of our proposed MALA* optimizer and alternative optimization strategies on the UGR and entropy.
An overview is shown in Table~\ref{tab:ablations}.
For each design choice, we report both the absolute and relative changes in entropy ($\Delta H$), unique grasp rate ($\Delta UGR$), and average max penetration depth ($\Delta D$), which are evaluated for the power grasp configuration and averaged across all hand types.
The results demonstrate that using MALA* consistently improves entropy (H) and unique grasp rate (UGR), and reduces penetration depth (D).
In order to motivate our design choices, we further investigate the influence of our optimization formulation and solver.
We show that reformulating the optimization problem as a nonlinear constraint problem (Sec.~\ref{sec:nonlinear_prob} using Theseus~\citep{pineda2022theseus}) leads to a significant degradation in performance, highlighting the necessity for hard constraints within the optimization formulation.
A similar (but less strong) decline is observed when reformulating the form closure condition from (iii) to (ii) (Theorem~\ref{thm:range} and additionally including the constraint $\sum_i^N{\alpha_i} = N$, to prevent the convergence to $\alpha_i \rightarrow 0$).

\begin{table}[t]
    \centering
    \caption{\textsc{\textbf{Ablation Study:}} Impact of optimizer variants and design choices on grasp quality. We report changes in entropy (H$\uparrow$), unique grasp rate (UGR$\uparrow$), and hand-object penetration (D$\downarrow$). Improvements from replacing MALA with our MALA$^*$ optimizer are shown across different methods, while subsequent rows analyze the effect of changing parts of the optimizer, optimization problem and relaxing the form closure condition in Theorem~\eqref{thm:range} from (iii) to (ii) \changed{or using the unconstrained softmax formulation $\alpha_i = \frac{\exp{\beta_i}}{\sum_j \exp(\beta_j)}$}. All values are calculated and averaged over all power grasps for all dexterous hands.}
    \resizebox{1.0\textwidth}{!}{
        \begin{tabular}{l|c|c|r|r|r} 
        \toprule  
        
        Method & Optimizer  & Energy Formulation  & $\Delta$ UGR $\uparrow$ \tiny{[\%]} & $\Delta$H $\uparrow$  & $\Delta$ D $ \downarrow$\tiny{[mm]} \\
        % Method & Change & $\Delta$ & $\Delta [\%] $ & $\Delta [mm]$  \\
            \midrule
        GenDexGrasp & \multirow{3}{*}{MALA $\rightarrow$ MALA$^*$} & Liu et al.~\citep{liu2021synthesizing} & \green{$+$7.0} & \green{$+$0.3}  & \green{-0.3} \\
        TDG & &  Chen et al.~\citep{chen2024task} & \green{$+$5.0}& \green{$+$0.2} & \green{-0.3} \\ 
        % \midrule
         \cmidrule{1-1}  \cmidrule{3-6}          
        \multirow{7}{*}{\OurMethod} &  & \multirow{3}{*}{ours} & \green{$+$5.4} &  \green{$+$0.2} &  \green{-0.4}\\
         \cmidrule{2-2}  \cmidrule{4-6}       
        
        % % \changed{
        & MALA$^*$ $\rightarrow$ MALA$^*$w/o resets $\texttt{\ \ \ \ \ }$& & \red{-4.8}  & \red{-0.1}& \red{$+$0.1}\\
        &MALA$^*$ $\rightarrow$ MALA$^*$w/o temp scaling & & \red{-4.5}  & \red{-0.1}  &  $+$0.0 \\
         \cmidrule{2-6}  
        &\multirow{4}{*}{MALA$^*$}   & ours $\rightarrow$ ours w/o exp& \red{-0.3}& $+$0.0  &\green{-0.1} \\
         & & Constrained (iii) $\rightarrow$ Unconstrained  $\texttt{\ \ \ \ \ \ \ }$ & \red{-10.0}  & \red{-0.2}  &\red{$+$0.1} \\
        
        % \OurMethod & QPTH $\rightarrow$QP & -7.6 / {-15\%} & -0.12  / {-4\%} & \\
        
        & & Constrained (iii) $\rightarrow$ Unconstrained-Softmax& \red{-10.1}  &\red{-0.3} & \green{-0.3}\\
       & & Constrained (iii) $\rightarrow$ Constrained (ii) $\texttt{\ \ \ \ \ \ }$&\red{-4.8} &\red{-0.1}  & \green{-0.2}\\
        \bottomrule 
        \end{tabular}}
    \label{tab:ablations}
    \vspace{-6mm}
\end{table}
% \
\section{Conclusion} 
\label{sec:conclusion}
In this work, we proposed a method to synthesize diverse and stable grasps for a wide range of objects and robotic grippers.
Our approach combines a differentiable force closure energy, defined through a Quadratic Program (QP), with an adjusted optimization strategy (MALA*) that promotes diversity during grasp synthesis.
We showed that our method is capable of generating a broad set of stable and unique grasps across different object geometries and gripper types, outperforming existing approaches in terms of both grasp stability and entropy.
Finally, we provide a detailed ablation study and analysis, validating the effectiveness of each component of our method.
As future work, we plan to extend our framework to dynamic manipulation scenarios and aim to use the generated dataset to train deep learning models for real-world grasp synthesis and robotic deployment.

\section{Limitations}
\label{sec:limitations}
While our method can generate a diverse set of stable grasps for a wide range of objects and grippers, there are still some limitations to our approach.
For the analytical optimization, our results demonstrate that incorporating a rigorous force closure metric directly enhances grasp quality and diversity.
However, this comes at the cost of increased computational complexity, with our optimization-based formulation requiring approximately 1.5--3 times more computation time than baseline methods (worst case \changed{3.4 s / grasp (ours) vs 1.15 s / grasp for the 24-DoF Shadow Hand}).
While this is acceptable for offline grasp synthesis, it may pose limitations for real-time applications or when used as a reward function in reinforcement learning.
A potential solution could involve implementing ADMM solvers~\citep{boyd2011distributed} on GPU.

Another challenge is the occurrence of mode collapse, where multiple grasps converge to very similar configurations, reducing the number of unique grasps.
Introducing coupling between grasp proposals to prevent them from occupying the same spatial region could help alleviate this issue.
Although preliminary experiments using density-based repulsion forces have not yielded promising results, we intend to explore this direction further in future work.
Finally, we observed some limitations within the Isaac Sim physics engine, where fingertips sometimes penetrate the object surface, leading to potential false positives in the grasp evaluation.

\section*{Acknowledgments}
This research was primarily supported by the ETH AI Center. This project received funding under the European Union’s Horizon Europe Framework Programme under grant agreement No 101121321, an ETH Zurich Research Grant No. 22-2 ETH-47 and the Swiss National Science Foundation through the National Centre of Competence in Automation (NCCR automation).
The authors thank Sabrina Bodmer for valuable discussions and for their help with proofreading and revising the manuscript.

\bibliography{main}  % .bib

\newpage
\section{Appendix}

\subsection{Metrics}
\label{app:entropy}

\emph{Entropy Formulation}:
 Directly calculating the entropy of the quaternion representation or the euler angles introduces bias due to the non-uniformity and/or singularities of the representation.
    We therefore calculate the entropy of grasp orientation by first converting it into axis-angle representation, which ensures that each orientation lies within a sphere of radius $\pi$.
    By finally converting the axis-angles into spherical coordinates, we can calculate the entropy of the orientation as the joint entropy of the spherical coordinates. 
    For each object, we calculate the joint entropy of the pose $\chi$ (position $\chi_p$, orientation $\chi_\theta$) and joint states $q$ and report the mean over all objects. The final entropy is calculated as 
    $$ H = \frac{1}{2}H(q) + \frac{1}{2}H(\chi) = \frac{1}{2}H(q) + \frac{1}{2}\left(H(\chi_p) + H(\chi_\theta)\right),$$ where $H(\bullet)$ denotes the entropy of the respective discretized distribution $\bullet$. We calculate the entropy by using uniform discretizations with 32 evenly spaced bins per dimension and assume each component to be independent.
    
\emph{Penetration Depth}: The penetration metric captures the amount of penetration between the object $O$ and the gripper $\mathcal{G}$. 
    It is calculated as the sum of the maximal penetration depths of a discrete set of points ($p_s(O) \in R^{3000 \times 3}$) uniformly sampled from the object surface. For each of these points, we calculate the max penetration depth with respect to all gripper links $d(\cdot, \cdot)$. The final penetration depth is the mean over all worst case penetrations, i.e.
    $$P =  \frac{1}{N_{\text{obj}} \times N_{\text{grasps}}} \sum_{(O, \mathcal{G})} \max_{{p \in p_s(O)}} (d(p, \mathcal{G})),$$
where $N_{\text{obj}}$ denotes the number of objects and $N_{\text{grasps}}$ the number of predicted grasps for each object (i.e. 32).

\emph{Success Rate with regard to applied Force:}\\
We employ two different success metrics \\
i) \emph{Succ}$^1$: As used in \citep{li2022gendexgrasp}, this denotes the success rate for withstanding a force applied along at least one of the main canonical axes (i.e., $\pm x$, $\pm y$, or $\pm z$). 
To ensure consistency with prior work, this metric is used by default to determine successful grasps (e.g., in calculating Unique Grasp Rate (UGR) or Entropy (H)), unless specified otherwise:
$$
\text{Succ}^1 = \frac{1}{N_{\text{obj}} \times N_{\text{grasps}}} \sum_{(O, \mathcal{G})} \mathbbm{1}\left[ \bigvee_{a \in \{\pm x, \pm y, \pm z\}} \text{Success}(O, \mathcal{G}, a) \right],
$$
where \text{Success} is true if the center of gravity of the object $O$ grasped by the grasp $\mathcal{G}$ remains stationary (within a radius of 3cm) during the application of the force along the axis $a$.\\
ii) \emph{Succ}$^3$: A more rigorous formulation of the success rate, reporting the proportion of cases where the grasp can withstand forces applied along all three main canonical axes:

$$
\text{Succ}^3 = \frac{1}{N_{\text{obj}} \times N_{\text{grasps}}} \sum_{(O, \mathcal{G})} \mathbbm{1} \left[ \bigwedge_{a \in \{\pm x, \pm y, \pm z\}} \text{Success}(O, \mathcal{G}, a) \right].
$$

% Where:

% * $N_{\text{obj}}$ is the number of objects,
% * $N_{\text{grasps}}$ is the number of grasp trials per object,
% * $\text{Success}(O, \mathcal{G}, a)$ is an indicator function returning 1 if the grasp $\mathcal{G}$ on object $O$ withstands force along axis $a$,
% * $\bigvee$ denotes a logical OR over the canonical axes,
% * $\mathbb{1}[\cdot]$ is the indicator function returning 1 if the condition inside is true, 0 otherwise.

\newpage
\subsection{Metropolis-adjusted Langevin algorithm (MALA)}
We base our implementation on the MALA optimizer used in \citep{wang2023dexgraspnet} and highlight our key differences (shown in blue) in the pseudo code \ref{alg:mala_optim}. We refer the reader to \citep{wang2023dexgraspnet} for the implementation of the full algorithm.

\begin{algorithm}[h!]
    \caption{Analytical Grasp Generation}
    \label{alg:mala_optim}
    \SetAlgoLined
    \KwIn{Initial grasp proposals $\mathcal{G}_{0} = \{ G^{(i)}_{0}\}_{i=0}^{N_{\text{grasps}}}$, Number of iterations $N$, temperature $T$.}
    \KwOut{Predicted Grasps $\mathcal{G}_N$}
    \For{$t=1$ to $N$}{
        $\hat{\mathcal{C}}' = ...$ \textcolor{gray}{// Re-sample active contact points} \\
        $(\hat \chi_t, \hat q_t) = ...$ \textcolor{gray}{// Calculate next grasp using SGD, based on energy} \\
        $ \hat{\mathcal{G}}_{t} = (\hat \chi_t, \hat q_t, \hat{\mathcal{C}}')$ \\

        % \tikzmk{A} $m_t = \beta_1 m_{t-1} + (1-\beta_1) \nabla E_t$ 
        % $v_t = \beta_2 v_{t-1} + (1-\beta_2) \nabla E_t^2$ 
        % $\hat{G}_t = G'_t - \gamma_t \hat{m_t} / (\sqrt{\hat{v_t}} + \epsilon)$ \tikzmk{B}
        % \boxit{myblue}
        % $\hat{G}_{x'} = \text{Resample}(\mathcal{X})$ 
        $\Delta E = E(\hat{\mathcal{G}}_{t}) - E(\mathcal{G}_{t-1})$  \\
        \textcolor{gray}{// Update According to Metropolis-Hastings} \\
        \textcolor{blue}{$\mu_t, \sigma_t = \text{mean}( E(\hat{\mathcal{G}}_{t})), \text{std}( E(\hat{\mathcal{G}}_{t}))$} 

        \For{$i=0$ to $N_{\text{grasps}}$}{
            \textcolor{blue}{$T^{(i)}_t = T_t \cdot \left(1 + \Phi(E(\hat{\mathcal{G}}_t^{(i)}); \mu_t, \sigma_t)\right)$}\\
            $p \sim \mathcal{U}(0,1)$ \\
            \If{$p < e^{-\Delta E^{(i)} / T_t^{(i)}}$}{
                $\mathcal{G}_{t}^{(i)} = \hat{\mathcal{G}}_{t}^{(i)}$  \textcolor{gray}{// Accept Sample}
            }\Else{
                $\mathcal{G}_{t}^{(i)} = \mathcal{G}_{t-1}^{(i)}$  \textcolor{gray}{// Reject Sample}
            }
        }
    }
\end{algorithm}

% \begin{table}[]

%     \centering
%     \begin{tabular}{l|cc}
%         \toprule
%         Method &
%         Success Rate $[\%] \uparrow$ & Avg 
%         Position Err. $[cm]\downarrow$\\
%         \midrule
%         \OurMethod & 70.4 $\pm$ 3& \textbf{28.3 }$\pm \textbf{3.0}$ \\
%         DexGraspNet & \textbf{70.5} $\pm$ 5 & 29.6 $\pm 6.0$ \\
%         TDG & 64.9 $\pm$ 6 & 34.3 $\pm 4.0$ \\
%         \bottomrule

%     \end{tabular}
%     \label{tab:my_label}
% \end{table}

\subsection{Implementation Details}
\label{app:impl_details}

\oursubsubsection{Simulation Environment}
We use Isaac Lab \citep{mittal2023orbit} (along with Isaac Sim) as our simulation environment, which offers high-fidelity physics simulations and a broad selection of robotic grippers and objects. The underlying physics time step is set to 200 Hz, and actuator commands are also applied at 200 Hz. We disable gravity for both the object and the gripper and configure the actuators torque and velocity limits based on realistic, gripper-specific values taken from their respective datasheets. A proportional-derivative (PD) controller with a high proportional gain (approximately 100) is employed, effectively mimicking a position controller for each gripper. The maximum simulation time is set to 4.8 seconds, with the external force changing every 0.5 seconds. For the Shadow Hand, we replace the tendon driven actuators with direct-drive actuators, as we observed instability when using tendons during simulation.

\oursubsubsection{Grasp Generation}
To allow for a fair comparison, we use the same hyperparameters and optimization setup as done in \citep{wang2023dexgraspnet}.
We modify the handcrafted set of contact points with contact meshes. We then randomly sample a discrete set of  contact points from these contact meshes using farthest point sampling to ensure good coverage.
These points are then used as potential contact points $\mathcal{C}$ for the optimization process.
For Pinch and Precision grasps, we select a subset of the contact meshes (i.e. fingertips of thumb, index for pinch, index  and middle for precision grasps). 

We reset all environments after $n_{reset}$ steps if the energy $E_t$ is below one standard deviation of the overall distribution (i.e., $p_{th} = 0.8413$).
Finally, we optimize over a total of $7000$ steps and evaluate the final grasp proposals using our simulation environment.

% value = (-h).relu()  + (- 1/t * torch.log(h.relu() + 1e-32)).relu()
In case of the unconstrained optimization problem, we use the implementation of Theseus \citep{pineda2022theseus} and unroll the last four optimization steps to compute the gradient. We implement the barrier functions as 
\begin{align*}
    \text{barrier}(\hat\gamma) = - h_{[h < 0]} - (1/\tau \cdot \log(h + \varepsilon))_{[h \geq 0]},
\end{align*}
with $\tau = 100$, $\epsilon = 10^{-6}$ and $h = [\hat\gamma-1, u - \hat\gamma]$.

\oursubsubsection{Dataset Processing}
We postprocess the data set introduced in \citep{wang2023dexgraspnet}. Besides re-meshing for better collision detection in Isaac Sim, we also re-scale the assets (which by default are normalized to the unit cube). To this end, we first scale each mesh by a factor of 0.08, which should make the assets roughly fit within an 8cm bounding box. However, we observe that not all meshes follow the unit-cube convention, meaning that a few become disproportionately small or large after this initial scaling. To address these issues, we compute the bounding box dimensions for each rescaled mesh and apply a corrective rescaling step: if the smallest axis exceeds 8cm, we downscale the mesh to ensure that the smallest axis is at most 8cm wide. If the largest axis falls below 7cm, we upscale it to ensure the the largest axis is at least 7cm large. This ensures that all assets remain within a consistent and physically meaningful size range.

\oursubsubsection{Baseline Methods}
\begin{itemize}
    \item DexGraspNet~\citep{wang2023dexgraspnet}:
    We use the official DexGraspNet implementation as the basis for our experiments. To improve performance, we adjust the relative weighting between force and torque by a factor of 5, as we found this leads to higher success rates. Specifically, we compute the energy as: 
    $$ E_{FC} = \left\|   G c \right\|_{W} \text{with}\ W = \begin{bmatrix} 1 & 0 \\ 0 & 5 \end{bmatrix}.$$
    \item GenDexGrasp~\citep{li2022gendexgrasp}: We implement GenDexGrasp by adding their normal based distance weighting.  Hence, the final energy formulation becomes:
\[
E_{\text{dis}}^{\texttt{GenDex}} = \sum_{c \in \mathcal{C}} e^{(1 - \langle n_c, n_o \rangle)} \| d(c, O) \|_2
\]
where $n_c$ is the normal vector at the fingertip, $n_o$ is the surface normal of the closest point on the object, and $d(c, O)$ denotes the closest distance from the contact point $c$ to the object surface $O$.
    
    \item MultiGripperDataset~\citep{casas2024multigrippergraspdatasetroboticgrasping}: We follow the same procedures as the authors and use the code provided by the authors for the supported grippers. More specifically, we use GraspIt! to sample an initial set of grasps using a maximal amount of $100000$ steps. Similarly to \citep{casas2024multigrippergraspdatasetroboticgrasping},we sample the final (in our case $32$) grasps using farthest point sampling. These grasps are then evaluated using our evaluation environment.
    \item TDG~\citep{chen2024task}: We use the implementation provided by the authors. We modify the relative force/torque weighting to $1/5$ and scale the energy term by a factor of 100 to make sure that the energy term has a comparable magnitude with the other baseline methods. Furthermore, we additionally add the normal weighting term from GenDexGrasp, as this has lead to improved performance. 
    \item GraspQP (ours): We use a friction coefficient of 0.2 and upper limit on $\hat\lambda$ of 50. We again, use a relative force/torque weighting of $1/5$ and use the normal based distance weighting from \citep{li2022gendexgrasp}.

\end{itemize}

\newpage

\subsection{Full Results}
\oursubsubsection{Numerical Results and Success Rates}
For completeness, we report the unique grasp rate and entropy (Tab. \ref{tab:main_results_ugr_h}) and grasp success rate (Tab. \ref{tab:main_results_succ_rate}) for all force thresholds, baselines and grasp types.

\label{app:full_results}

\begin{table*}[h!]
    \caption{
		extsc{Unique Grasp Rate (UGR) and Entropy for Various Interaction Forces}: We list the UGR and entropy (H) for various interaction forces, ranging from 1 to 10 N.
	Succ$^3$ denotes the success rate to withstand all interaction forces along all three main directions ($\pm x$, $\pm y$, $\pm z$), and Succ$^1$ denotes the success rate along either one of the main directions.
    % \todo{Check DexGraspNet}
    }
\label{tab:main_results_ugr_h}
\setlength{\tabcolsep}{3pt}
\centering
        
\resizebox{.9\textwidth}{!}{
	\begin{tabular}{l|l|l|cc|cc|cc|cc|cc||cc|c}
		\toprule
		        
		 Grasp  & \multirow{ 2}{*}{Method}  &  \multirow{ 2}{*}{Optimizer} &  \multicolumn{2}{c|}{Allegro} &  \multicolumn{2}{c|}{Shadow Hand}&  \multicolumn{2}{c|}{Robotiq3f} &  \multicolumn{2}{c|}{Robotiq2f} &  \multicolumn{2}{c|}{Ability Hand}  & \multicolumn{2}{c}{Overall} \\
		    
		Type  &   & &UGR $\uparrow$ 
		%& SR$\uparrow$ 
		&   H  $\uparrow$ 
		&   UGR $\uparrow$  
		% & SR$\uparrow$ 
		&   H $\uparrow$ 
		& UGR $\uparrow$ 
		% & SR $\uparrow$ 
		&   H $\uparrow$ 
		&  UGR $\uparrow$ 
		% & SR$\uparrow$ 
		&   H $\uparrow$ 
		&  UGR $\uparrow$ 
		% & SR$\uparrow$ 
		&   H$\uparrow$ 
		&  UGR $\uparrow$ 
		% & SR$\uparrow$ 
		&   H $\uparrow$ \\
		\midrule  
 \textbf{1N} \\
		\midrule  
 
\multirow[c]{5}{*}{power} & \multirow[c]{2}{*}{DexGraspNet} & MALA & 68\% & 3.54 & 43\% & 3.12 & 50\% & 3.08 & \underline{32\%} & 3.28 & 20\% & 1.99 & 43\% & 3.00 \\
 &  & MALA$^*$ & 69\% & 3.51 & 49\% & 3.23 & 61\% & 3.27 & \textbf{33\%} & 3.28 & 20\% & 1.86 & 46\% & 3.03 \\

 & GenDexGrasp & MALA$^*$ & \underline{74\%} & \underline{3.56} & \underline{53\%} & \underline{3.34} & 73\% & \underline{3.36} & 32\% & \textbf{3.40} & 24\% & 2.22 & \underline{51\%} & \underline{3.18} \\
 & TDG & MALA$^*$ & 69\% & 3.51 & 53\% & 3.34 & \underline{77\%} & 3.31 & 29\% & \underline{3.30} & \underline{27\%} & \underline{2.36} & 51\% & 3.16 \\

 & GraspQP (ours) & MALA$^*$ & \textbf{77\%} & \textbf{3.61} & \textbf{67\%} & \textbf{3.56} & \textbf{78\%} & \textbf{3.39} & 29\% & 3.23 & \textbf{37\%} & \textbf{2.73} & \textbf{58\%} & \textbf{3.30} \\

 \midrule

\multirow[c]{5}{*}{pinch} & \multirow[c]{2}{*}{DexGraspNet} & MALA & 32\% & 2.52 & 29\% & 2.50 & 17\% & 1.93 & & \textbf{--} & 15\% & 1.73 & 23\% & 2.17 \\
 &  & MALA$^*$ & 36\% & 2.69 & 26\% & 2.27 & 22\% & 2.10 & & \underline{--} & 13\% & 1.51 & 24\% & 2.14 \\

 & GenDexGrasp & MALA$^*$ & \underline{44\%} & 2.96 & \underline{37\%} & \underline{2.77} & \underline{40\%} & \textbf{2.90} & & -- & \underline{23\%} & \underline{2.18} & \underline{36\%} & \underline{2.70} \\
 & TDG & MALA$^*$ & 43\% & \underline{2.99} & 33\% & 2.61 & 38\% & 2.79 & & -- & 21\% & 1.99 & 34\% & 2.59 \\

 & GraspQP (ours) & MALA$^*$ & \textbf{50\%} & \textbf{3.11} & \textbf{40\%} & \textbf{2.87} & \textbf{41\%} & \underline{2.89} & & -- & \textbf{29\%} & \textbf{2.35} & \textbf{40\%} & \textbf{2.81} \\

 \midrule

\multirow[c]{5}{*}{precision} & \multirow[c]{2}{*}{DexGraspNet} & MALA & 39\% & 2.82 & 35\% & 2.60 & 23\% & 2.26 & & \textbf{--} & 17\% & 1.82 & 29\% & 2.38 \\
 &  & MALA$^*$ & 43\% & 2.94 & 32\% & 2.48 & 31\% & 2.54 & & \underline{--} & 16\% & 1.63 & 30\% & 2.40 \\

 & GenDexGrasp & MALA$^*$ & 48\% & 3.11 & 42\% & 2.94 & 45\% & \underline{3.08} & & -- & \underline{21\%} & \underline{1.98} & \underline{39\%} & 2.78 \\
 & TDG & MALA$^*$ & \underline{49\%} & \underline{3.27} & \underline{43\%} & \underline{3.01} & \underline{47\%} & 3.07 & & -- & 19\% & 1.85 & 39\% & \underline{2.80} \\

 & GraspQP (ours) & MALA$^*$ & \textbf{57\%} & \textbf{3.36} & \textbf{48\%} & \textbf{3.07} & \textbf{49\%} & \textbf{3.11} & & -- & \textbf{32\%} & \textbf{2.53} & \textbf{46\%} & \textbf{3.02} \\

\midrule
\textbf{2N} \\
\midrule
\multirow[c]{5}{*}{power} & \multirow[c]{2}{*}{DexGraspNet} & MALA & 67\% & 3.52 & 42\% & 3.09 & 45\% & 2.97 & \underline{32\%} & 3.27 & 20\% & 1.90 & 41\% & 2.95 \\
 &  & MALA$^*$ & 68\% & 3.51 & 47\% & 3.22 & 55\% & 3.13 & \textbf{33\%} & 3.27 & 19\% & 1.84 & 44\% & 2.99 \\

 & GenDexGrasp & MALA$^*$ & \underline{75\%} & \underline{3.56} & \underline{52\%} & 3.31 & 69\% & \textbf{3.30} & 32\% & \textbf{3.40} & 24\% & 2.29 & \underline{50\%} & \underline{3.17} \\
 & TDG & MALA$^*$ & 68\% & 3.51 & 52\% & \underline{3.33} & \textbf{72\%} & 3.25 & 29\% & \underline{3.29} & \underline{27\%} & \underline{2.33} & 49\% & 3.14 \\

 & GraspQP (ours) & MALA$^*$ & \textbf{76\%} & \textbf{3.60} & \textbf{67\%} & \textbf{3.55} & \underline{71\%} & \underline{3.26} & 29\% & 3.23 & \textbf{37\%} & \textbf{2.70} & \textbf{56\%} & \textbf{3.27} \\

 \midrule

\multirow[c]{5}{*}{pinch} & \multirow[c]{2}{*}{DexGraspNet} & MALA & 32\% & 2.52 & 23\% & 2.23 & 9\% & 1.08 & & \textbf{--} & 13\% & 1.56 & 19\% & 1.85 \\
 &  & MALA$^*$ & 36\% & 2.67 & 19\% & 1.95 & 11\% & 1.27 & & \underline{--} & 12\% & 1.34 & 20\% & 1.81 \\

 & GenDexGrasp & MALA$^*$ & \underline{44\%} & 2.94 & \underline{30\%} & \underline{2.54} & 17\% & 1.77 & & -- & \underline{21\%} & \underline{2.06} & \underline{28\%} & \underline{2.33} \\
 & TDG & MALA$^*$ & 43\% & \underline{2.99} & 28\% & 2.41 & \textbf{20\%} & \underline{1.96} & & -- & 19\% & 1.87 & 27\% & 2.31 \\

 & GraspQP (ours) & MALA$^*$ & \textbf{50\%} & \textbf{3.11} & \textbf{34\%} & \textbf{2.65} & \underline{19\%} & \textbf{1.96} & & -- & \textbf{27\%} & \textbf{2.24} & \textbf{32\%} & \textbf{2.49} \\

 \midrule

\multirow[c]{5}{*}{precision} & \multirow[c]{2}{*}{DexGraspNet} & MALA & 39\% & 2.80 & 31\% & 2.49 & 14\% & 1.59 & & \textbf{--} & 17\% & 1.75 & 25\% & 2.15 \\
 &  & MALA$^*$ & 42\% & 2.91 & 28\% & 2.35 & 18\% & 1.84 & & \underline{--} & 14\% & 1.52 & 26\% & 2.15 \\

 & GenDexGrasp & MALA$^*$ & 49\% & 3.12 & 38\% & 2.81 & 26\% & 2.38 & & -- & \underline{21\%} & \underline{1.98} & \underline{34\%} & 2.57 \\
 & TDG & MALA$^*$ & \underline{50\%} & \underline{3.27} & \underline{39\%} & \underline{2.90} & \underline{28\%} & \textbf{2.41} & & -- & 18\% & 1.78 & 34\% & \underline{2.59} \\

 & GraspQP (ours) & MALA$^*$ & \textbf{56\%} & \textbf{3.35} & \textbf{44\%} & \textbf{2.99} & \textbf{28\%} & \underline{2.39} & & -- & \textbf{30\%} & \textbf{2.39} & \textbf{40\%} & \textbf{2.78} \\

 \midrule
 \textbf{5N} \\
 \midrule
\multirow[c]{5}{*}{power} & \multirow[c]{2}{*}{DexGraspNet} & MALA & 66\% & 3.51 & 35\% & 2.86 & 33\% & 2.56 & 30\% & 3.21 & 18\% & 1.77 & 37\% & 2.78 \\
 &  & MALA$^*$ & 68\% & 3.51 & 41\% & 3.04 & 40\% & 2.76 & \underline{31\%} & 3.23 & 18\% & 1.72 & 40\% & 2.85 \\

 & GenDexGrasp & MALA$^*$ & \underline{74\%} & \underline{3.56} & \underline{47\%} & 3.16 & 51\% & \underline{2.93} & \textbf{32\%} & \textbf{3.38} & 23\% & \underline{2.19} & \underline{45\%} & \underline{3.04} \\
 & TDG & MALA$^*$ & 69\% & 3.51 & 47\% & \underline{3.19} & \underline{54\%} & 2.93 & 28\% & \underline{3.28} & \underline{25\%} & 2.16 & 45\% & 3.01 \\

 & GraspQP (ours) & MALA$^*$ & \textbf{76\%} & \textbf{3.60} & \textbf{63\%} & \textbf{3.46} & \textbf{57\%} & \textbf{3.02} & 29\% & 3.22 & \textbf{35\%} & \textbf{2.63} & \textbf{52\%} & \textbf{3.19} \\

 \midrule

\multirow[c]{5}{*}{pinch} & \multirow[c]{2}{*}{DexGraspNet} & MALA & 31\% & 2.47 & 10\% & 1.28 & \underline{4\%} & \underline{0.49} & & \textbf{--} & 11\% & 1.25 & 14\% & 1.37 \\
 &  & MALA$^*$ & 35\% & 2.64 & 7\% & 0.92 & 3\% & 0.35 & & \underline{--} & 10\% & 1.09 & 14\% & 1.25 \\

 & GenDexGrasp & MALA$^*$ & \underline{44\%} & 2.92 & \underline{14\%} & \underline{1.57} & 4\% & 0.39 & & -- & \underline{17\%} & \underline{1.78} & \underline{20\%} & 1.66 \\
 & TDG & MALA$^*$ & 42\% & \underline{2.96} & 13\% & 1.43 & \textbf{5\%} & \textbf{0.58} & & -- & 16\% & 1.72 & 19\% & \underline{1.67} \\

 & GraspQP (ours) & MALA$^*$ & \textbf{49\%} & \textbf{3.11} & \textbf{17\%} & \textbf{1.84} & 4\% & 0.45 & & -- & \textbf{23\%} & \textbf{2.09} & \textbf{23\%} & \textbf{1.87} \\

 \midrule

\multirow[c]{5}{*}{precision} & \multirow[c]{2}{*}{DexGraspNet} & MALA & 37\% & 2.73 & 14\% & 1.64 & 5\% & 0.55 & & \textbf{--} & 14\% & 1.53 & 18\% & 1.61 \\
 &  & MALA$^*$ & 43\% & 2.94 & 14\% & 1.52 & 6\% & 0.65 & & \underline{--} & 11\% & 1.32 & 18\% & 1.61 \\

 & GenDexGrasp & MALA$^*$ & 48\% & 3.06 & \underline{23\%} & 2.18 & \textbf{9\%} & \underline{1.07} & & -- & \underline{18\%} & \underline{1.80} & \underline{25\%} & \underline{2.02} \\
 & TDG & MALA$^*$ & \underline{49\%} & \underline{3.26} & 23\% & \underline{2.30} & 8\% & 0.87 & & -- & 16\% & 1.59 & 24\% & 2.00 \\

 & GraspQP (ours) & MALA$^*$ & \textbf{55\%} & \textbf{3.32} & \textbf{29\%} & \textbf{2.45} & \underline{9\%} & \textbf{1.14} & & -- & \textbf{26\%} & \textbf{2.28} & \textbf{30\%} & \textbf{2.30} \\

 \midrule
 \textbf{10N} \\
 \midrule
 
\multirow[c]{5}{*}{power} & \multirow[c]{2}{*}{DexGraspNet} & MALA & 65\% & 3.49 & 23\% & 2.31 & 22\% & 2.01 & 29\% & 3.16 & 17\% & 1.65 & 31\% & 2.53 \\
 &  & MALA$^*$ & 67\% & 3.50 & 28\% & 2.55 & 28\% & 2.31 & \textbf{31\%} & 3.20 & 17\% & 1.66 & 34\% & 2.64 \\

 & GenDexGrasp & MALA$^*$ & \underline{72\%} & \underline{3.53} & 35\% & \underline{2.83} & 37\% & \underline{2.60} & \underline{31\%} & \textbf{3.34} & 21\% & 2.01 & \underline{39\%} & \underline{2.86} \\
 & TDG & MALA$^*$ & 67\% & 3.49 & \underline{36\%} & 2.79 & \underline{42\%} & 2.58 & 28\% & \underline{3.24} & \underline{23\%} & \underline{2.07} & 39\% & 2.83 \\

 & GraspQP (ours) & MALA$^*$ & \textbf{76\%} & \textbf{3.59} & \textbf{48\%} & \textbf{3.12} & \textbf{45\%} & \textbf{2.75} & 28\% & 3.21 & \textbf{33\%} & \textbf{2.56} & \textbf{46\%} & \textbf{3.05} \\

 \midrule

\multirow[c]{5}{*}{pinch} & \multirow[c]{2}{*}{DexGraspNet} & MALA & 30\% & 2.45 & \underline{2\%} & 0.16 & \textbf{2\%} & \textbf{0.22} & & \textbf{--} & 8\% & 0.92 & 11\% & 0.94 \\
 &  & MALA$^*$ & 34\% & 2.58 & 1\% & 0.02 & 1\% & 0.05 & & \underline{--} & 7\% & 0.80 & 11\% & 0.86 \\

 & GenDexGrasp & MALA$^*$ & \underline{42\%} & 2.89 & 2\% & \underline{0.25} & 1\% & 0.14 & & -- & \underline{14\%} & \underline{1.61} & \underline{15\%} & \underline{1.22} \\
 & TDG & MALA$^*$ & 42\% & \underline{2.94} & \textbf{3\%} & \textbf{0.35} & 2\% & \underline{0.18} & & -- & 12\% & 1.25 & 15\% & 1.18 \\

 & GraspQP (ours) & MALA$^*$ & \textbf{48\%} & \textbf{3.08} & 2\% & 0.23 & \underline{2\%} & 0.14 & & -- & \textbf{19\%} & \textbf{1.80} & \textbf{18\%} & \textbf{1.31} \\

 \midrule

\multirow[c]{5}{*}{precision} & \multirow[c]{2}{*}{DexGraspNet} & MALA & 37\% & 2.72 & 3\% & 0.24 & 2\% & 0.20 & & \textbf{--} & 11\% & 1.27 & 13\% & 1.11 \\
 &  & MALA$^*$ & 41\% & 2.91 & 2\% & 0.15 & 2\% & 0.19 & & \underline{--} & 10\% & 1.12 & 14\% & 1.09 \\

 & GenDexGrasp & MALA$^*$ & 46\% & 3.05 & 5\% & 0.72 & \textbf{3\%} & \textbf{0.33} & & -- & \underline{15\%} & \underline{1.49} & \underline{17\%} & \underline{1.40} \\
 & TDG & MALA$^*$ & \underline{48\%} & \underline{3.21} & \underline{6\%} & \underline{0.73} & 2\% & 0.19 & & -- & 12\% & 1.37 & 17\% & 1.38 \\

 & GraspQP (ours) & MALA$^*$ & \textbf{54\%} & \textbf{3.30} & \textbf{8\%} & \textbf{0.91} & \underline{3\%} & \underline{0.27} & & -- & \textbf{22\%} & \textbf{2.02} & \textbf{22\%} & \textbf{1.63} \\

		\bottomrule
	\end{tabular} 
}
    \vspace{-5mm}
\end{table*}
\begin{table*}[h!]
	\caption{{\textsc{Grasp Success Rates for Various Interaction Forces}}: We list the raw success rates for various interaction forces, ranging from 1 to 10 N.
	Succ$^3$ denotes the success rate to withstand all interaction forces along all three main directions ($\pm x$, $\pm y$, $\pm z$), and Succ$^1$ denotes the success rate along either one of the main directions.
    % \todo{Check DexGraspNet}
    }
\label{tab:main_results_succ_rate}
\setlength{\tabcolsep}{3pt}
\centering
        
\resizebox{\textwidth}{!}{
	\begin{tabular}{l|l|l|cc|cc|cc|cc|cc||cc|c}
		\toprule
		        
		 Grasp  & \multirow{ 2}{*}{Method}  &  \multirow{ 2}{*}{Optimizer} &  \multicolumn{2}{c|}{Allegro} &  \multicolumn{2}{c|}{Shadow Hand}&  \multicolumn{2}{c|}{Robotiq3f} &  \multicolumn{2}{c|}{Robotiq2f} &  \multicolumn{2}{c|}{Ability Hand}  & \multicolumn{2}{c}{Overall} \\
		    
		Type  &   & &   Succ$^1$  $\uparrow$ & Succ$^3$ $\uparrow$ 
		%& SR$\uparrow$ 
		&   Succ$^1$  $\uparrow$ 
        & Succ$^3$ $\uparrow$ 
		%& SR$\uparrow$ 
		&   Succ$^1$  $\uparrow$ & Succ$^3$ $\uparrow$ 
		%& SR$\uparrow$ 
		&   Succ$^1$  $\uparrow$ & Succ$^3$ $\uparrow$ 
		%& SR$\uparrow$ 
		&   Succ$^1$  $\uparrow$ & Succ$^3$ $\uparrow$ 
		%& SR$\uparrow$ 
		&   Succ$^1$  $\uparrow$ & Succ$^3$ $\uparrow$ 
		%& SR$\uparrow$ 
		 \\
		\midrule  
 \textbf{1N} \\
		\midrule  
        
\multirow[c]{5}{*}{power} & \multirow[c]{2}{*}{DexGraspNet} & MALA & 71\% & 50\% & 45\% & 37\% & 53\% & 34\% & 64\% & 58\% & 21\% & 12\% & 51\% & 38\% \\
 &  & MALA$^*$ & 71\% & 48\% & 50\% & 41\% & 64\% & 42\% & 70\% & 66\% & 20\% & 10\% & 55\% & 41\% \\

 & GenDexGrasp & MALA$^*$ & \underline{77\%} & \underline{55\%} & \underline{55\%} & 47\% & 76\% & 54\% & 83\% & 76\% & 25\% & 14\% & 63\% & \underline{49\%} \\
 & TDG & MALA$^*$ & 71\% & 48\% & 55\% & \underline{49\%} & \underline{79\%} & \underline{55\%} & \underline{86\%} & \underline{79\%} & \underline{28\%} & \underline{15\%} & \underline{64\%} & 49\% \\

 & GraspQP (ours) & MALA$^*$ & \textbf{80\%} & \textbf{57\%} & \textbf{70\%} & \textbf{63\%} & \textbf{80\%} & \textbf{56\%} & \textbf{88\%} & \textbf{82\%} & \textbf{39\%} & \textbf{24\%} & \textbf{71\%} & \textbf{56\%} \\

 \midrule

\multirow[c]{5}{*}{pinch} & \multirow[c]{2}{*}{DexGraspNet} & MALA & 33\% & 27\% & 31\% & 19\% & 19\% & 5\% &  &  & 16\% & 10\% & 25\% & 15\% \\
 &  & MALA$^*$ & 38\% & 31\% & 28\% & 16\% & 23\% & 6\% &  &  & 14\% & 9\% & 26\% & 16\% \\

 & GenDexGrasp & MALA$^*$ & \underline{46\%} & \underline{41\%} & \underline{39\%} & \underline{26\%} & \underline{43\%} & \underline{10\%} & & & \underline{24\%} & \underline{16\%} & \underline{38\%} & \underline{23\%} \\
 & TDG & MALA$^*$ & 45\% & 40\% & 35\% & 23\% & 41\% & \textbf{12\%} & & & 22\% & 15\% & 36\% & 22\% \\

 & GraspQP (ours) & MALA$^*$ & \textbf{52\%} & \textbf{47\%} & \textbf{42\%} & \textbf{29\%} & \textbf{44\%} & 10\% & & & \textbf{30\%} & \textbf{21\%} & \textbf{42\%} & \textbf{27\%} \\

 \midrule

\multirow[c]{5}{*}{precision} & \multirow[c]{2}{*}{DexGraspNet} & MALA & 41\% & 32\% & 37\% & 26\% & 24\% & 9\% &  &  & 18\% & 11\% & 30\% & 20\% \\
 &  & MALA$^*$ & 44\% & 36\% & 34\% & 24\% & 33\% & 10\% &  &  & 16\% & 11\% & 32\% & 20\% \\

 & GenDexGrasp & MALA$^*$ & 50\% & 41\% & \underline{45\%} & \underline{34\%} & 46\% & 15\% & & & \underline{23\%} & \underline{15\%} & 41\% & 26\% \\
 & TDG & MALA$^*$ & \underline{51\%} & \underline{42\%} & 45\% & 34\% & \underline{49\%} & \textbf{20\%} & & & 20\% & 13\% & \underline{41\%} & \underline{27\%} \\

 & GraspQP (ours) & MALA$^*$ & \textbf{59\%} & \textbf{49\%} & \textbf{50\%} & \textbf{39\%} & \textbf{51\%} & \underline{19\%} & & & \textbf{33\%} & \textbf{22\%} & \textbf{48\%} & \textbf{32\%} \\

 \midrule
 \textbf{2N} \\
 \midrule
\multirow[c]{5}{*}{power} & \multirow[c]{2}{*}{DexGraspNet} & MALA & 69\% & 50\% & 44\% & 33\% & 47\% & 22\% & 63\% & 57\% & 21\% & 11\% & 49\% & 35\% \\
 &  & MALA$^*$ & 71\% & 50\% & 49\% & 37\% & 57\% & 28\% & 71\% & 64\% & 20\% & 10\% & 54\% & 38\% \\

 & GenDexGrasp & MALA$^*$ & \underline{77\%} & \underline{54\%} & \underline{54\%} & 43\% & 71\% & 37\% & 83\% & 76\% & 25\% & \underline{14\%} & 62\% & \underline{45\%} \\
 & TDG & MALA$^*$ & 70\% & 49\% & 54\% & \underline{44\%} & \textbf{74\%} & \underline{39\%} & \underline{85\%} & \underline{78\%} & \underline{28\%} & 14\% & \underline{62\%} & 45\% \\

 & GraspQP (ours) & MALA$^*$ & \textbf{78\%} & \textbf{56\%} & \textbf{69\%} & \textbf{59\%} & \underline{73\%} & \textbf{41\%} & \textbf{88\%} & \textbf{82\%} & \textbf{38\%} & \textbf{23\%} & \textbf{69\%} & \textbf{52\%} \\

 \midrule

\multirow[c]{5}{*}{pinch} & \multirow[c]{2}{*}{DexGraspNet} & MALA & 33\% & 27\% & 25\% & 10\% & 9\% & \textbf{2\%} &  &  & 14\% & 8\% & 20\% & 12\% \\
 &  & MALA$^*$ & 37\% & 30\% & 21\% & 7\% & 11\% & 1\% &  &  & 13\% & 7\% & 20\% & 11\% \\

 & GenDexGrasp & MALA$^*$ & \underline{46\%} & \underline{40\%} & \underline{32\%} & \underline{13\%} & 18\% & 1\% & & & \underline{22\%} & \underline{13\%} & \underline{30\%} & \underline{17\%} \\
 & TDG & MALA$^*$ & 44\% & 39\% & 30\% & 12\% & \textbf{21\%} & 2\% & & & 20\% & 13\% & 29\% & 16\% \\

 & GraspQP (ours) & MALA$^*$ & \textbf{52\%} & \textbf{47\%} & \textbf{36\%} & \textbf{16\%} & \underline{20\%} & \underline{2\%} & & & \textbf{28\%} & \textbf{18\%} & \textbf{34\%} & \textbf{21\%} \\

 \midrule

\multirow[c]{5}{*}{precision} & \multirow[c]{2}{*}{DexGraspNet} & MALA & 40\% & 32\% & 32\% & 15\% & 15\% & \underline{3\%} &  &  & 18\% & 10\% & 26\% & 15\% \\
 &  & MALA$^*$ & 44\% & 36\% & 30\% & 14\% & 18\% & 3\% &  &  & 15\% & 10\% & 27\% & 16\% \\

 & GenDexGrasp & MALA$^*$ & 51\% & 40\% & 40\% & 23\% & 27\% & 3\% & & & \underline{23\%} & \underline{13\%} & 35\% & 20\% \\
 & TDG & MALA$^*$ & \underline{52\%} & \underline{42\%} & \underline{41\%} & \underline{25\%} & \underline{29\%} & \textbf{4\%} & & & 20\% & 12\% & \underline{36\%} & \underline{21\%} \\

 & GraspQP (ours) & MALA$^*$ & \textbf{58\%} & \textbf{48\%} & \textbf{47\%} & \textbf{28\%} & \textbf{30\%} & 3\% & & & \textbf{31\%} & \textbf{20\%} & \textbf{42\%} & \textbf{25\%} \\

 \midrule
\textbf{5N} \\
 \midrule
\multirow[c]{5}{*}{power} & \multirow[c]{2}{*}{DexGraspNet} & MALA & 68\% & 49\% & 37\% & 18\% & 35\% & 8\% & 61\% & 54\% & 19\% & 9\% & 44\% & 28\% \\
 &  & MALA$^*$ & 71\% & 48\% & 42\% & 22\% & 42\% & 10\% & 69\% & 62\% & 19\% & 9\% & 49\% & 30\% \\

 & GenDexGrasp & MALA$^*$ & \underline{76\%} & \underline{53\%} & 48\% & 28\% & 52\% & 13\% & 83\% & 73\% & 24\% & 12\% & 57\% & 36\% \\
 & TDG & MALA$^*$ & 71\% & 49\% & \underline{49\%} & \underline{30\%} & \underline{55\%} & \underline{14\%} & \underline{85\%} & \underline{77\%} & \underline{25\%} & \underline{14\%} & \underline{57\%} & \underline{37\%} \\

 & GraspQP (ours) & MALA$^*$ & \textbf{79\%} & \textbf{55\%} & \textbf{65\%} & \textbf{40\%} & \textbf{59\%} & \textbf{16\%} & \textbf{88\%} & \textbf{81\%} & \textbf{36\%} & \textbf{22\%} & \textbf{65\%} & \textbf{43\%} \\

 \midrule

\multirow[c]{5}{*}{pinch} & \multirow[c]{2}{*}{DexGraspNet} & MALA & 32\% & 24\% & 11\% & 2\% & \underline{4\%} & \textbf{0\%} &  &  & 11\% & 6\% & 14\% & 8\% \\
 &  & MALA$^*$ & 37\% & 30\% & 8\% & 1\% & 3\% & \underline{0\%} &  &  & 11\% & 5\% & 15\% & 9\% \\

 & GenDexGrasp & MALA$^*$ & \underline{46\%} & \underline{39\%} & \underline{15\%} & 2\% & 4\% & 0\% & & & \underline{18\%} & 7\% & \underline{21\%} & 12\% \\
 & TDG & MALA$^*$ & 44\% & 38\% & 14\% & \underline{3\%} & \textbf{5\%} & 0\% & & & 17\% & \underline{8\%} & 20\% & \underline{12\%} \\

 & GraspQP (ours) & MALA$^*$ & \textbf{51\%} & \textbf{45\%} & \textbf{19\%} & \textbf{3\%} & 4\% & 0\% & & & \textbf{24\%} & \textbf{12\%} & \textbf{24\%} & \textbf{15\%} \\

 \midrule

\multirow[c]{5}{*}{precision} & \multirow[c]{2}{*}{DexGraspNet} & MALA & 39\% & 30\% & 15\% & 3\% & 5\% & 0\% &  &  & 15\% & 7\% & 19\% & 10\% \\
 &  & MALA$^*$ & 44\% & 35\% & 14\% & 2\% & 6\% & \textbf{1\%} &  &  & 12\% & 6\% & 19\% & 11\% \\

 & GenDexGrasp & MALA$^*$ & 50\% & 39\% & \underline{24\%} & 4\% & \underline{9\%} & 0\% & & & \underline{20\%} & \underline{10\%} & \underline{26\%} & 13\% \\
 & TDG & MALA$^*$ & \underline{51\%} & \underline{41\%} & 24\% & \underline{6\%} & 8\% & \underline{1\%} & & & 17\% & 9\% & 25\% & \underline{14\%} \\

 & GraspQP (ours) & MALA$^*$ & \textbf{57\%} & \textbf{47\%} & \textbf{31\%} & \textbf{8\%} & \textbf{10\%} & 0\% & & & \textbf{28\%} & \textbf{15\%} & \textbf{32\%} & \textbf{18\%} \\

 \midrule
 \textbf{10N} \\
 \midrule
\multirow[c]{5}{*}{power} & \multirow[c]{2}{*}{DexGraspNet} & MALA & 67\% & 47\% & 24\% & 7\% & 22\% & 2\% & 60\% & 49\% & 17\% & 8\% & 38\% & 23\% \\
 &  & MALA$^*$ & 70\% & 46\% & 29\% & 9\% & 29\% & 2\% & 68\% & 57\% & 18\% & 8\% & 43\% & 24\% \\

 & GenDexGrasp & MALA$^*$ & \underline{75\%} & \underline{51\%} & 35\% & 12\% & 38\% & \underline{4\%} & 81\% & 69\% & 22\% & 10\% & 50\% & 29\% \\
 & TDG & MALA$^*$ & 69\% & 47\% & \underline{37\%} & \underline{13\%} & \underline{43\%} & 3\% & \underline{84\%} & \underline{72\%} & \underline{24\%} & \underline{13\%} & \underline{51\%} & \underline{30\%} \\

 & GraspQP (ours) & MALA$^*$ & \textbf{78\%} & \textbf{54\%} & \textbf{50\%} & \textbf{20\%} & \textbf{47\%} & \textbf{5\%} & \textbf{87\%} & \textbf{76\%} & \textbf{34\%} & \textbf{21\%} & \textbf{59\%} & \textbf{35\%} \\

 \midrule

\multirow[c]{5}{*}{pinch} & \multirow[c]{2}{*}{DexGraspNet} & MALA & 31\% & 22\% & 2\% & \textbf{0\%} & \textbf{2\%} & \textbf{0\%} &  &  & 8\% & 3\% & 11\% & 6\% \\
 &  & MALA$^*$ & 36\% & 27\% & 1\% & \underline{0\%} & 1\% & \underline{0\%} &  &  & 8\% & 2\% & 12\% & 7\% \\

 & GenDexGrasp & MALA$^*$ & \underline{44\%} & \underline{37\%} & 2\% & 0\% & \underline{2\%} & 0\% & & & \underline{15\%} & 5\% & \underline{16\%} & \underline{10\%} \\
 & TDG & MALA$^*$ & 43\% & 34\% & \underline{3\%} & 0\% & 2\% & 0\% & & & 12\% & \underline{6\%} & 15\% & 10\% \\

 & GraspQP (ours) & MALA$^*$ & \textbf{50\%} & \textbf{41\%} & \textbf{3\%} & 0\% & 2\% & 0\% & & & \textbf{19\%} & \textbf{6\%} & \textbf{18\%} & \textbf{12\%} \\

 \midrule

\multirow[c]{5}{*}{precision} & \multirow[c]{2}{*}{DexGraspNet} & MALA & 39\% & 30\% & 3\% & \underline{0\%} & 2\% & \textbf{0\%} &  &  & 12\% & 4\% & 14\% & 8\% \\
 &  & MALA$^*$ & 43\% & 33\% & 2\% & 0\% & 2\% & \underline{0\%} &  &  & 10\% & 4\% & 14\% & 9\% \\

 & GenDexGrasp & MALA$^*$ & 48\% & 37\% & \underline{6\%} & 0\% & \textbf{3\%} & 0\% & & & \underline{16\%} & \underline{7\%} & \underline{18\%} & 11\% \\
 & TDG & MALA$^*$ & \underline{50\%} & \underline{39\%} & 6\% & 0\% & 2\% & 0\% & & & 14\% & 6\% & 18\% & \underline{11\%} \\

 & GraspQP (ours) & MALA$^*$ & \textbf{57\%} & \textbf{45\%} & \textbf{8\%} & \textbf{1\%} & \underline{3\%} & 0\% & & & \textbf{24\%} & \textbf{11\%} & \textbf{23\%} & \textbf{14\%} \\

		\bottomrule
	\end{tabular} 
}
    \vspace{-5mm}
\end{table*}

\vfill\newpage
\clearpage
\newpage
\subsubsection{Contact Maps} We additionally provide heatmaps that visualize the contact regions for each gripper and grasp type. To generate these maps, we compute the distance from each object vertex $v \in V(O)$ to the gripper surface for every grasp pose $G \in \mathcal{G}$. The color intensity $c(v)$ at each vertex is then calculated as the normalized sum of the exponentiated negative distances across all grasp poses:
$c(v) = \frac{1}{|\mathcal{G}|} \sum_{G \in \mathcal{G}} \left( \frac{\exp\left(-10 \cdot d(v, G)\right)}{\sum_{\vartheta \in V(O)} \exp\left(-10 \cdot d(\vartheta, G)\right)} \right)$.
Finally, the resulting intensities are visualized using the viridis colormap, where high-intensity regions appear yellow and low-intensity regions appear blue.

\begin{figure}[h]
 \centering
    \begin{subfigure}[b]{0.99\textwidth}        
        \includegraphics[trim={4cm 6cm 7cm 8cm},clip, width=\linewidth]{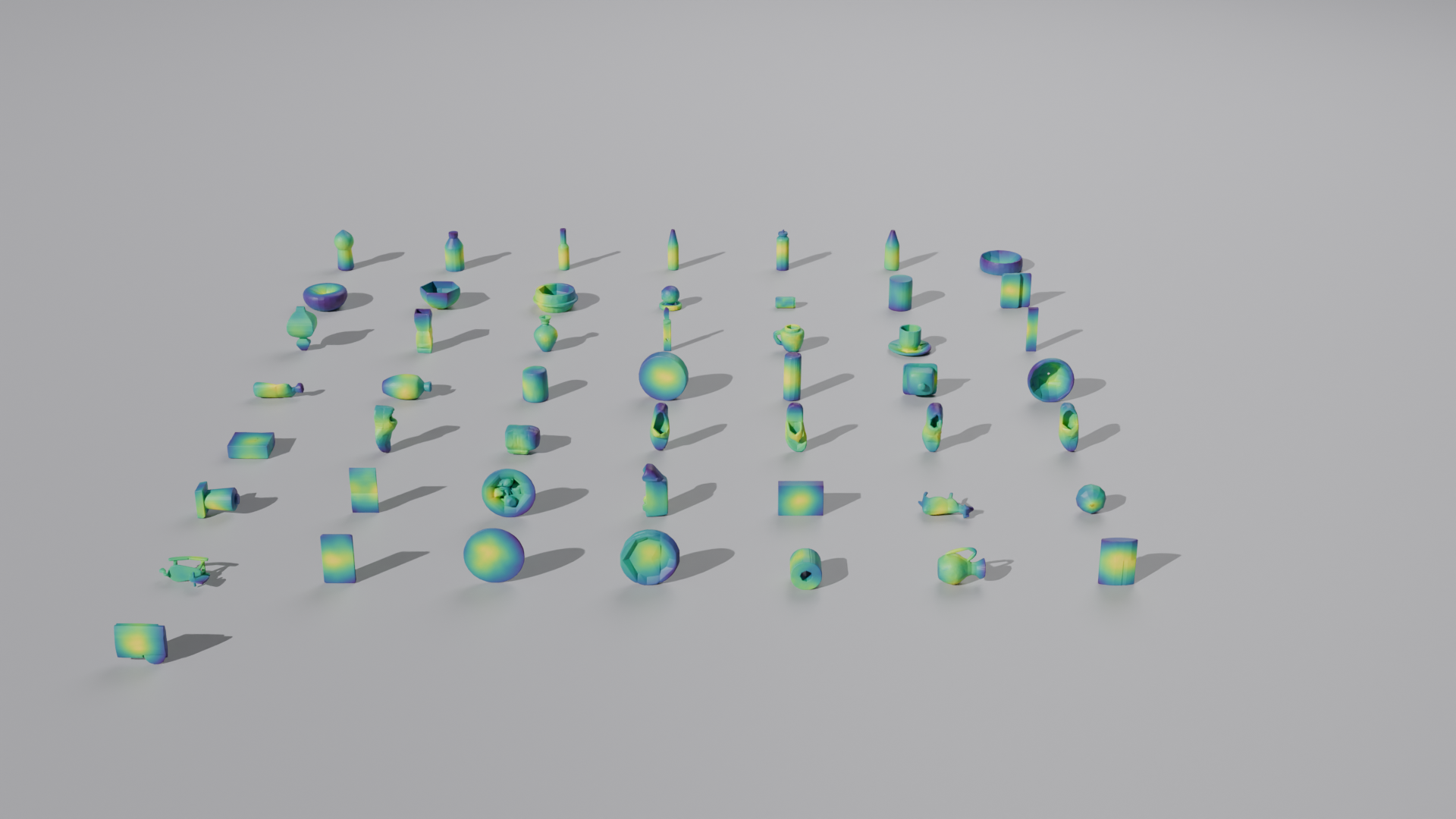}
        \caption{Ability Hand - default}
        %\label{fig:image1}
    \end{subfigure}
    \vspace{0.5cm}  % space between rows
    
    \begin{subfigure}[b]{0.49\textwidth}        
    \includegraphics[trim={4cm 6cm 7cm 8cm},clip, width=\linewidth]{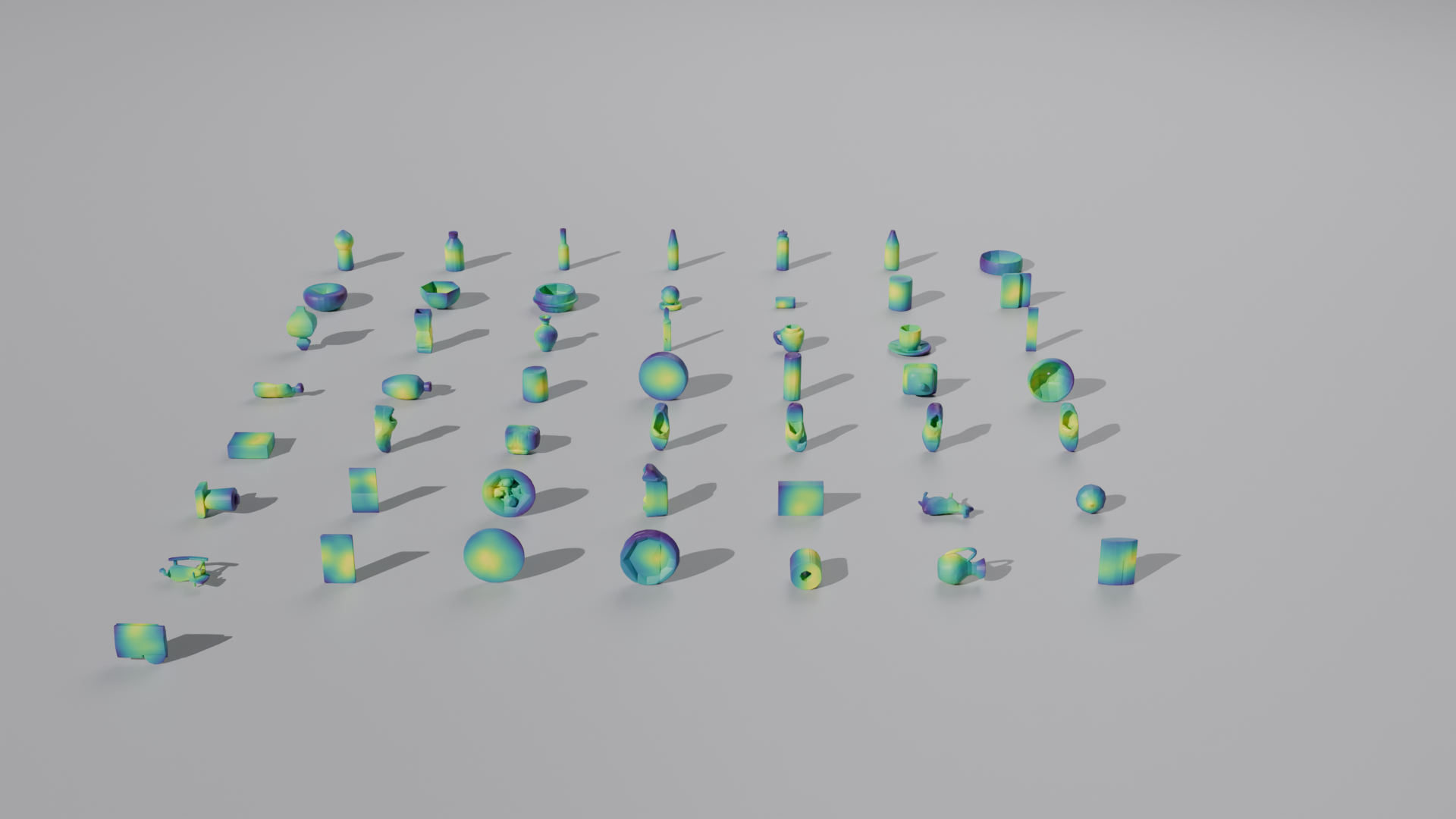}
        \caption{Ability Hand - precision}
        %\label{fig:image2}
    \end{subfigure}
    \hfill
    \begin{subfigure}[b]{0.49\textwidth}
        \includegraphics[trim={4cm 6cm 7cm 8cm},clip, width=\linewidth]{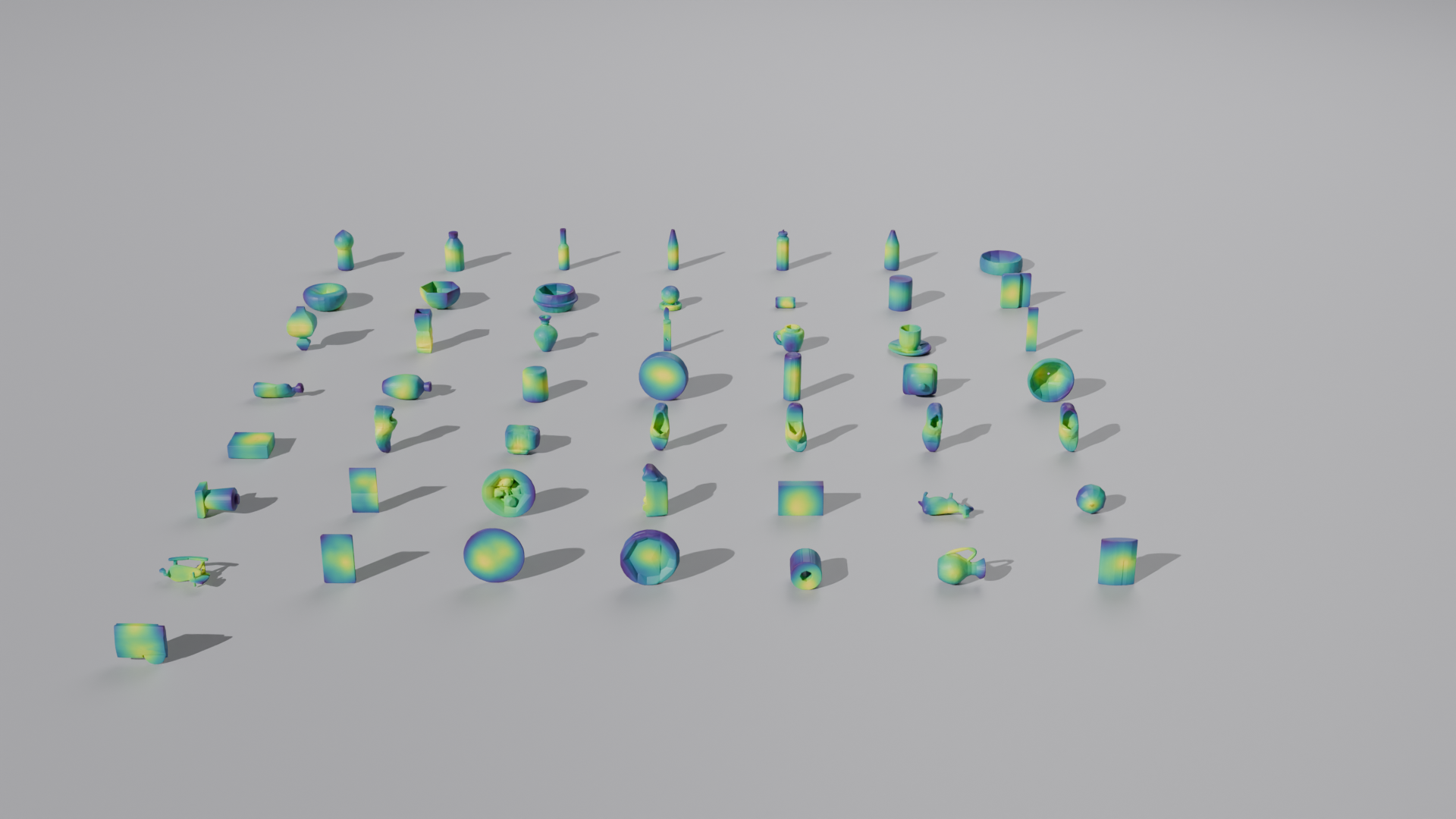}
        \caption{Ability Hand - pinch}
        %\label{fig:image1}
    \end{subfigure}
    \hfill
    \caption{
    Contact heatmaps for various objects using the Ability Hand. The color intensity (mapped with the viridis colormap) indicates regions of frequent contact across all grasp poses: yellow represents high-contact areas, while blue indicates low-contact regions. Notably, many objects show concentrated contact regions near their center of gravity.
    }
\end{figure}
\begin{figure}[h]
 \centering
    
    \begin{subfigure}[b]{0.99\textwidth}        
    \includegraphics[trim={4cm 6cm 7cm 8cm},clip, width=\linewidth]{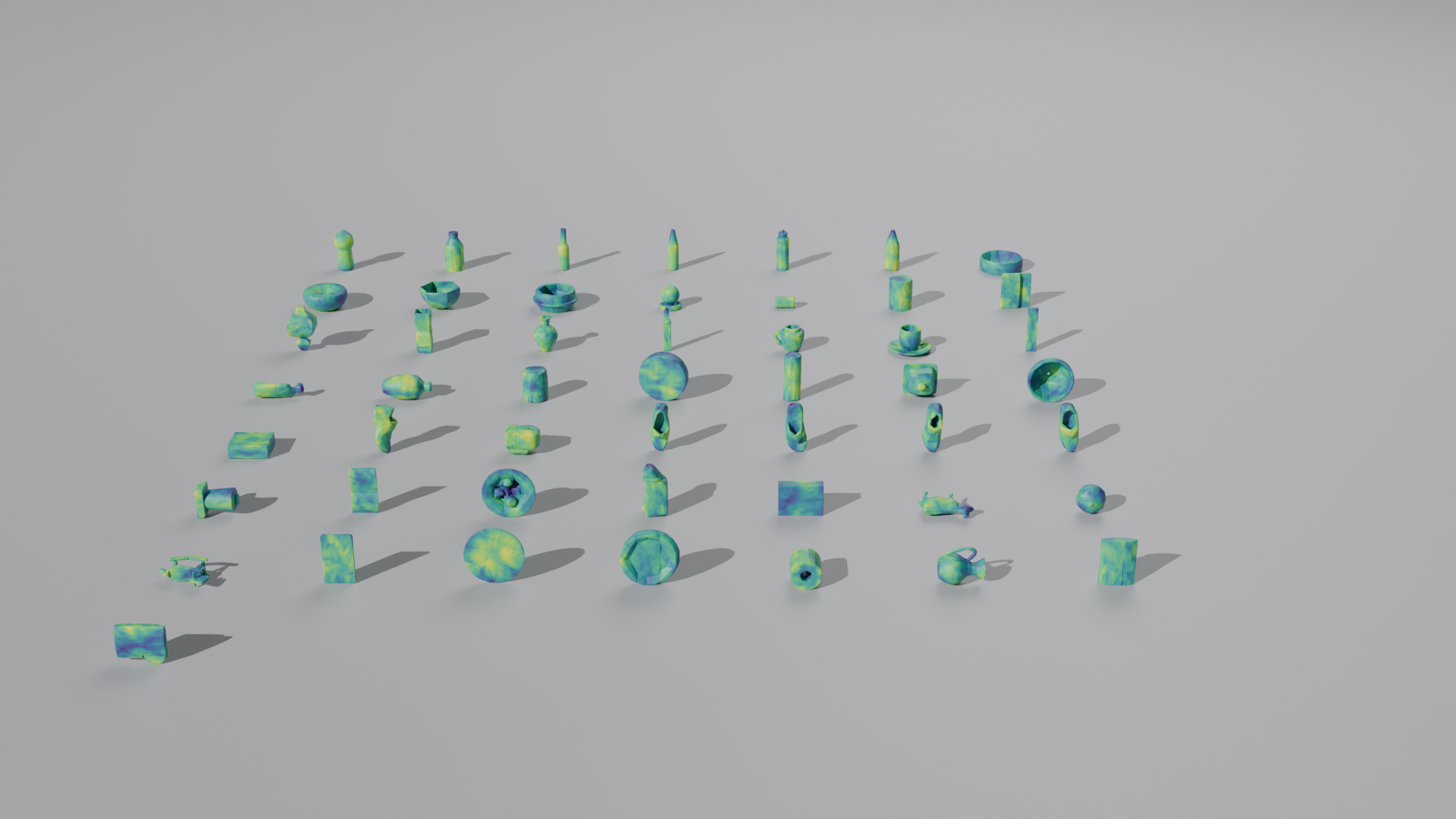}
        \caption{Roboti 3F - default}
        %\label{fig:image2}
    \end{subfigure}
    \vspace{0.5cm}  % space between rows
    
    \begin{subfigure}[b]{0.49\textwidth}
        \includegraphics[trim={4cm 6cm 7cm 8cm},clip, width=\linewidth]{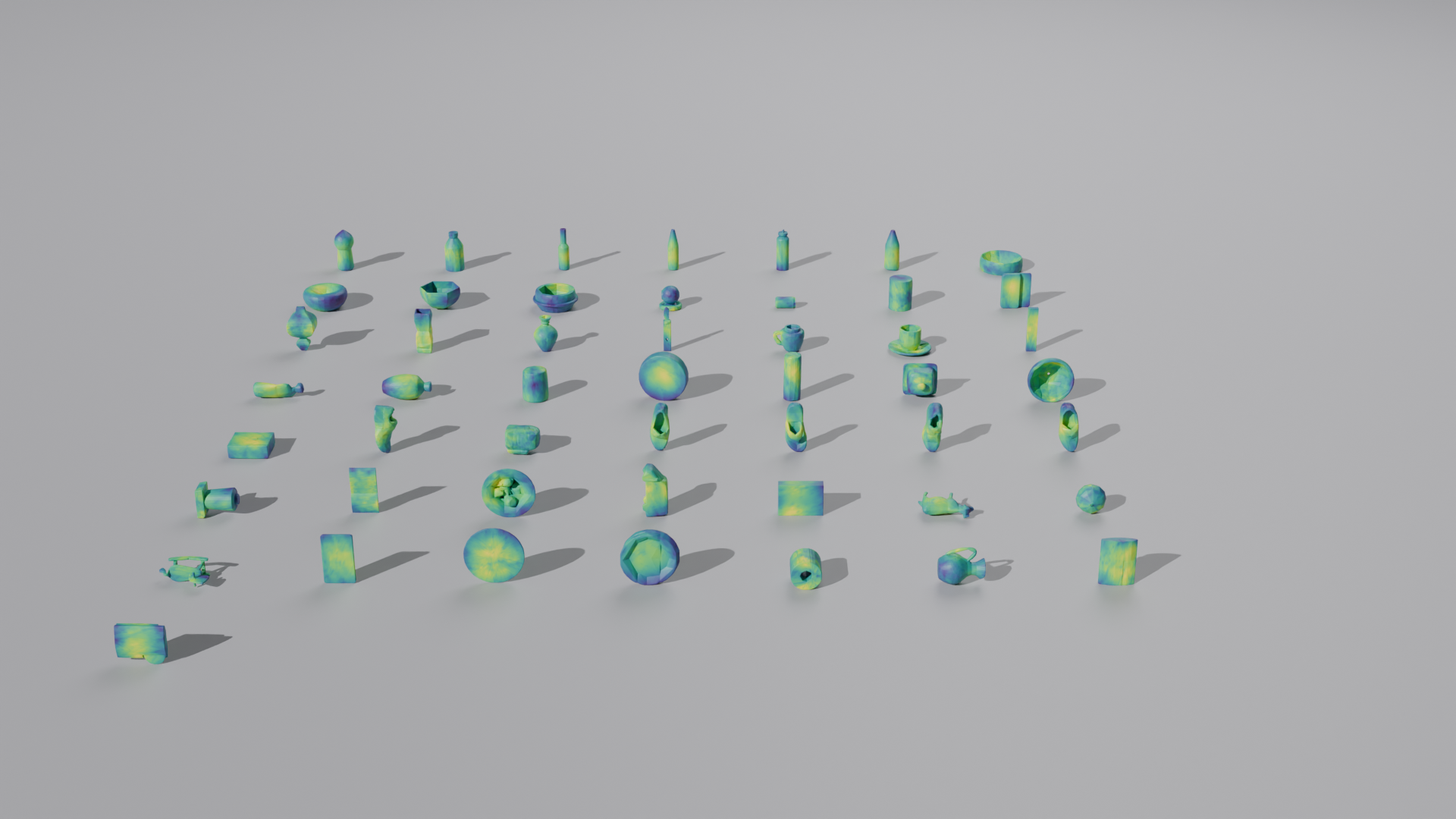}
        \caption{Roboti 3F - precision}
        %\label{fig:image1}
    \end{subfigure}
    \hfill
    \begin{subfigure}[b]{0.49\textwidth}        
    \includegraphics[trim={4cm 6cm 7cm 8cm},clip, width=\linewidth]{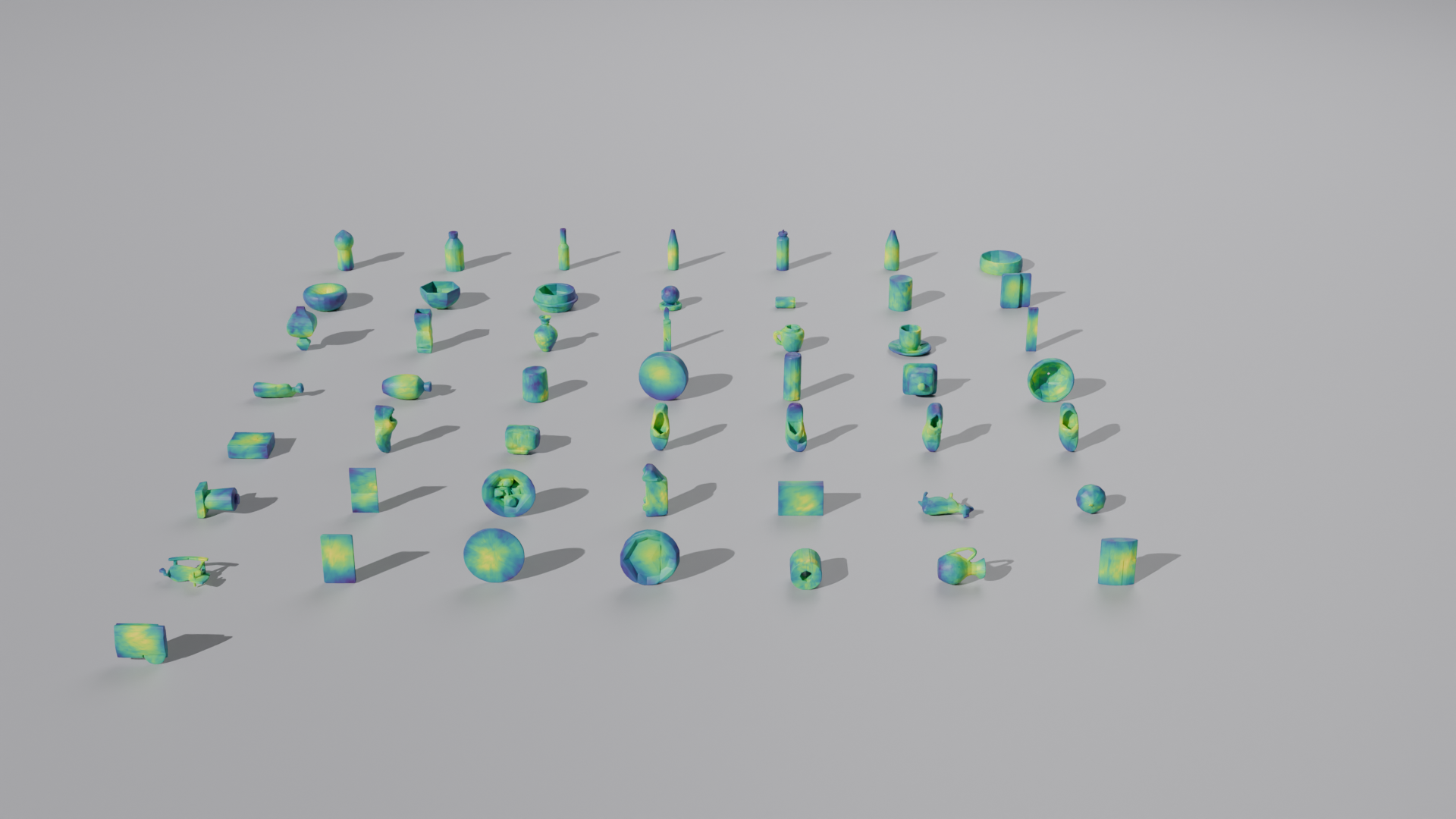}
        \caption{Roboti 3F - pinch}
        %\label{fig:image2}
    \end{subfigure}
    \vspace{0.5cm}  % space between rows
    
    \caption{
    Contact heatmaps for various objects using the Robotiq 3F. The color intensity (mapped with the viridis colormap) indicates regions of frequent contact across all grasp poses: yellow represents high-contact areas, while blue indicates low-contact regions.
    }
\end{figure}
\begin{figure}[h]
 \centering

    \begin{subfigure}[b]{0.96\textwidth}
        \includegraphics[trim={4cm 6cm 7cm 8cm},clip, width=\linewidth]{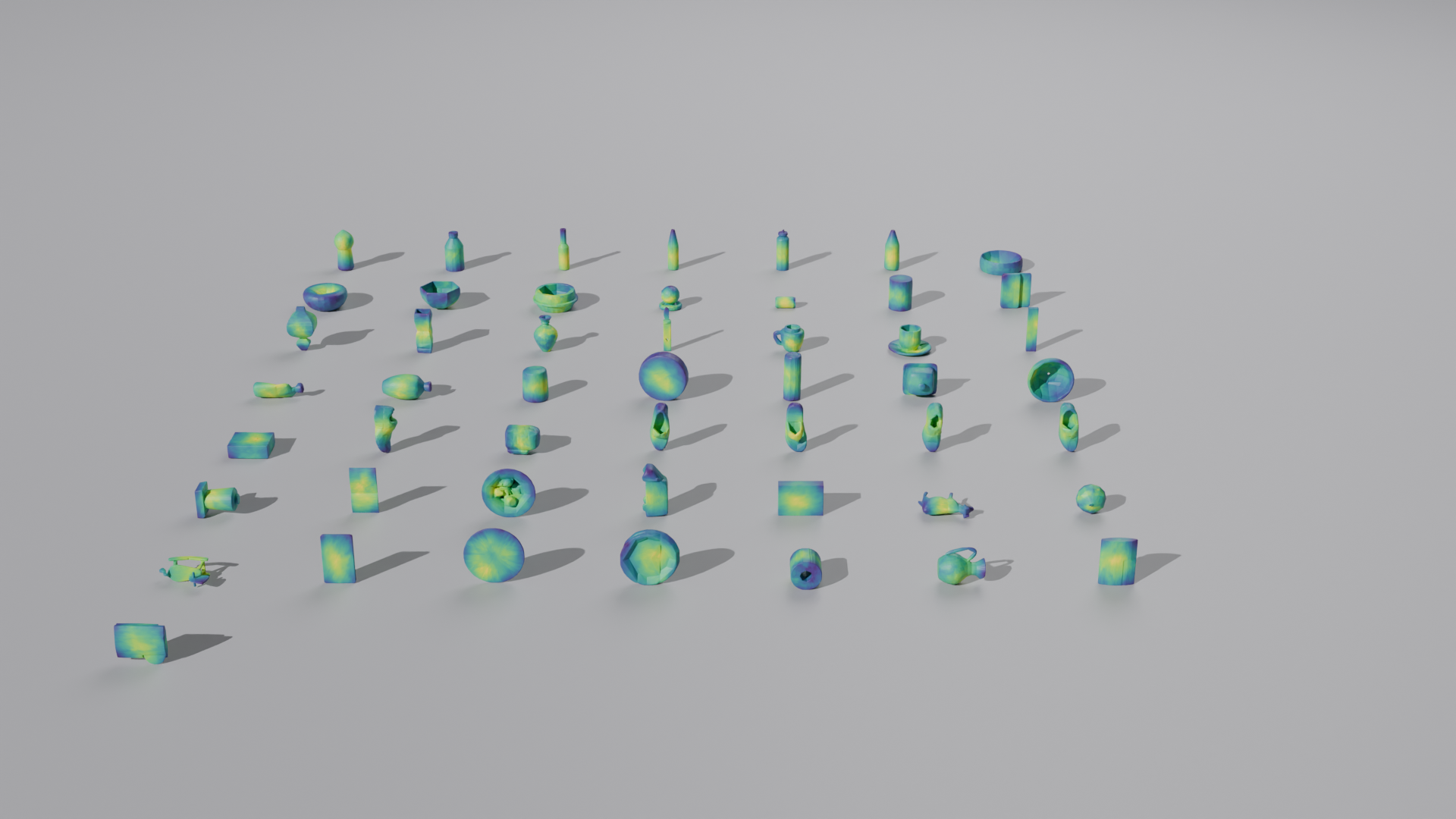}
        \caption{Shadow Hand- default}
        %\label{fig:image1}
        \vspace{0.5cm}  % space between rows
    \end{subfigure}
    \begin{subfigure}[b]{0.49\textwidth}        
    \includegraphics[trim={4cm 6cm 7cm 8cm},clip, width=\linewidth]{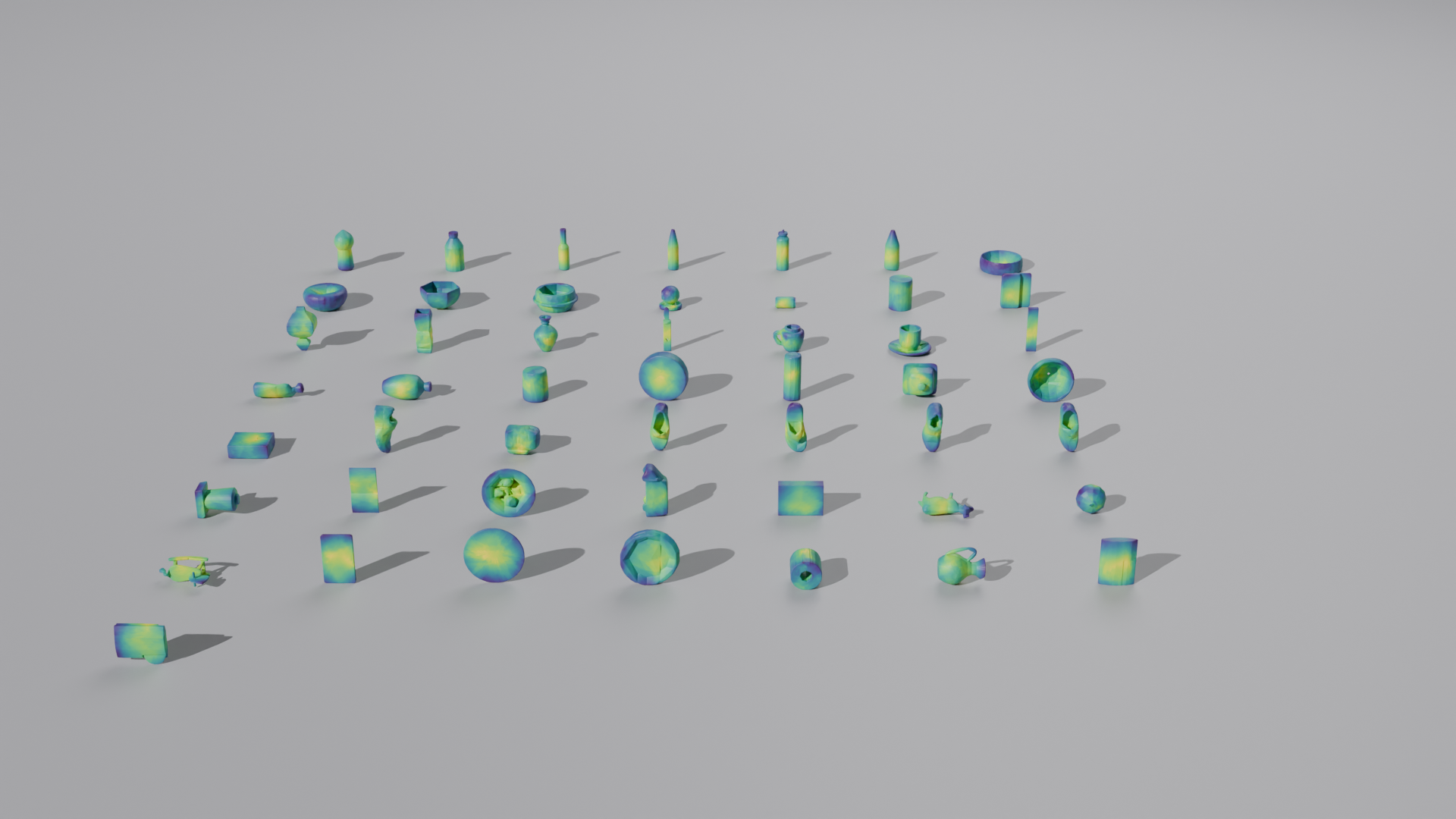}
        \caption{Shadow Hand- precision}
        %\label{fig:image2}
    \end{subfigure}
    \hfill
    \begin{subfigure}[b]{0.49\textwidth}
        \includegraphics[trim={4cm 6cm 7cm 8cm},clip, width=\linewidth]{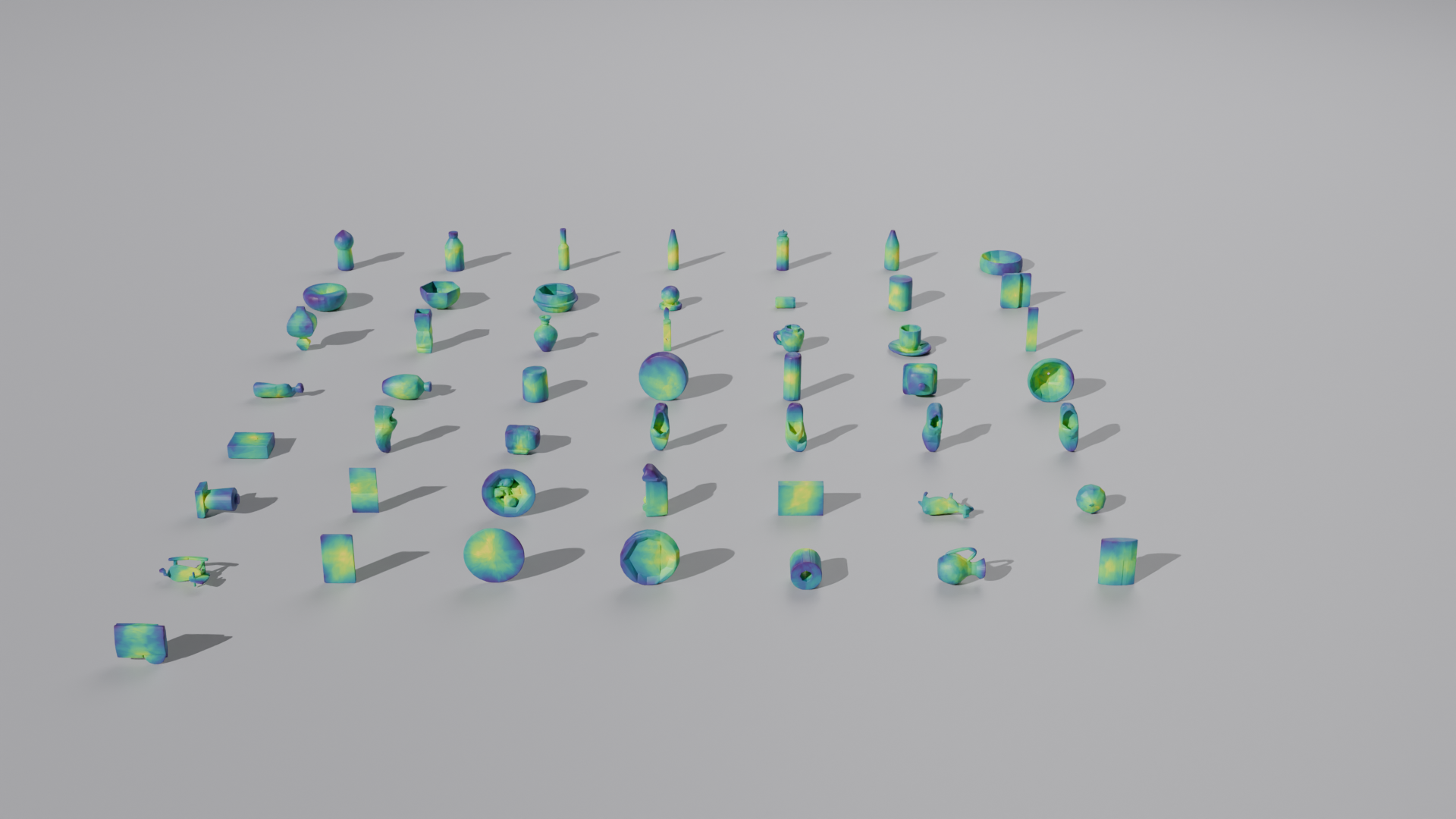}
        \caption{Shadow Hand- pinch}
        %\label{fig:image1}
    \end{subfigure}
    \hfill
    
    \caption{
    Contact heatmaps for various objects using the Shadow Hand. The color intensity (mapped with the viridis colormap) indicates regions of frequent contact across all grasp poses: yellow represents high-contact areas, while blue indicates low-contact regions. Notably, many objects show concentrated contact regions near their center of gravity.
    }
\end{figure}
\begin{figure}[h]
 \centering

    % \begin{subfigure}[b]{0.98\textwidth}        
    \includegraphics[trim={4cm 6cm 7cm 8cm},clip, width=\linewidth]{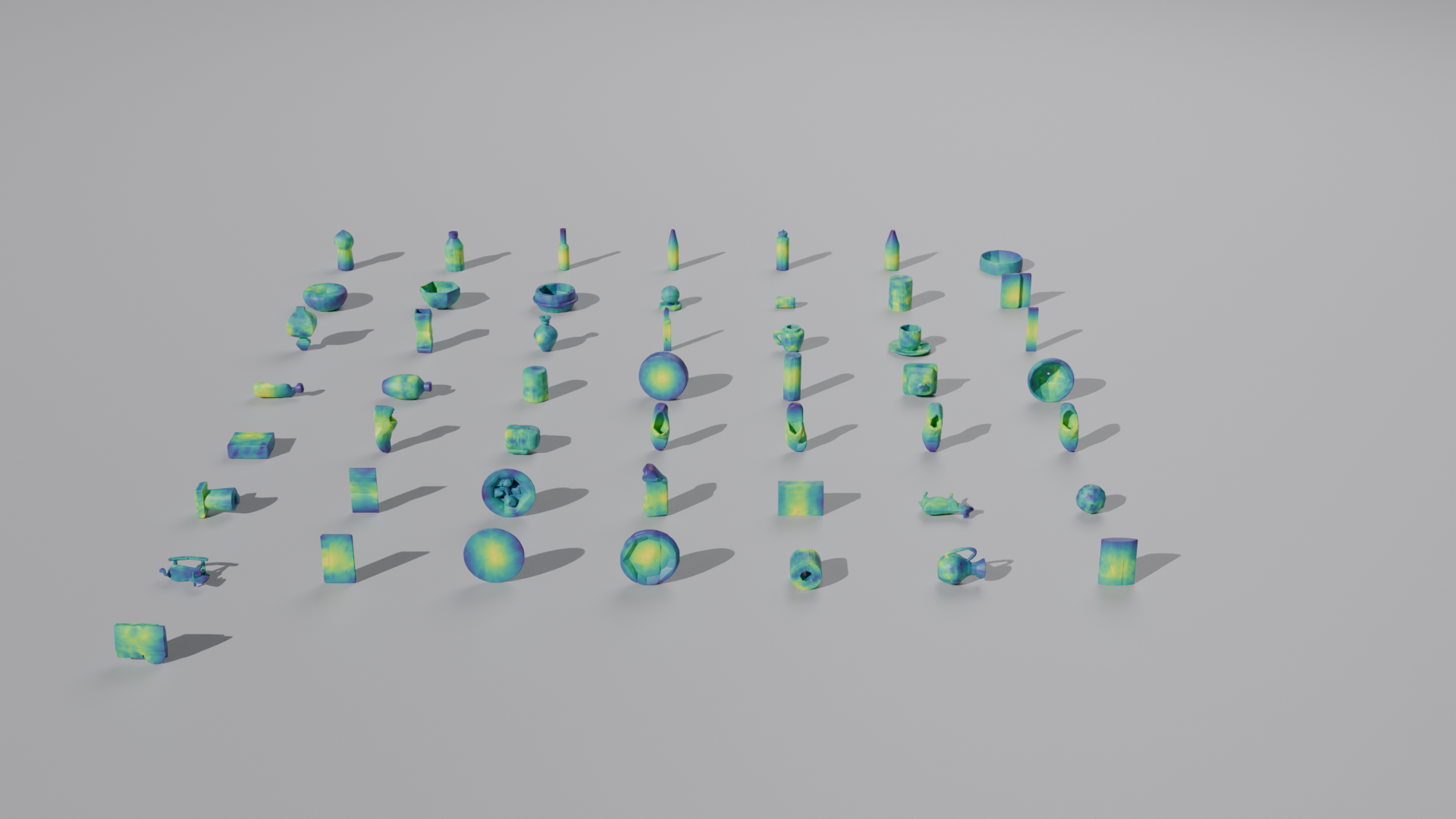}
        \caption{Robotiq 2f140}
        %\label{fig:image2}
    % \end{subfigure}
    
    \caption{
    Contact heatmaps for various objects using the Robotiq 2f140 Gripper. The color intensity (mapped with the viridis colormap) indicates regions of frequent contact across all grasp poses: yellow represents high-contact areas, while blue indicates low-contact regions. Notably, many objects show concentrated contact regions near their center of gravity.
    }
\end{figure}
\begin{figure}[h]
 \centering
    \begin{subfigure}[b]{0.98\textwidth}        
        \includegraphics[trim={4cm 6cm 7cm 8cm},clip, width=\linewidth]{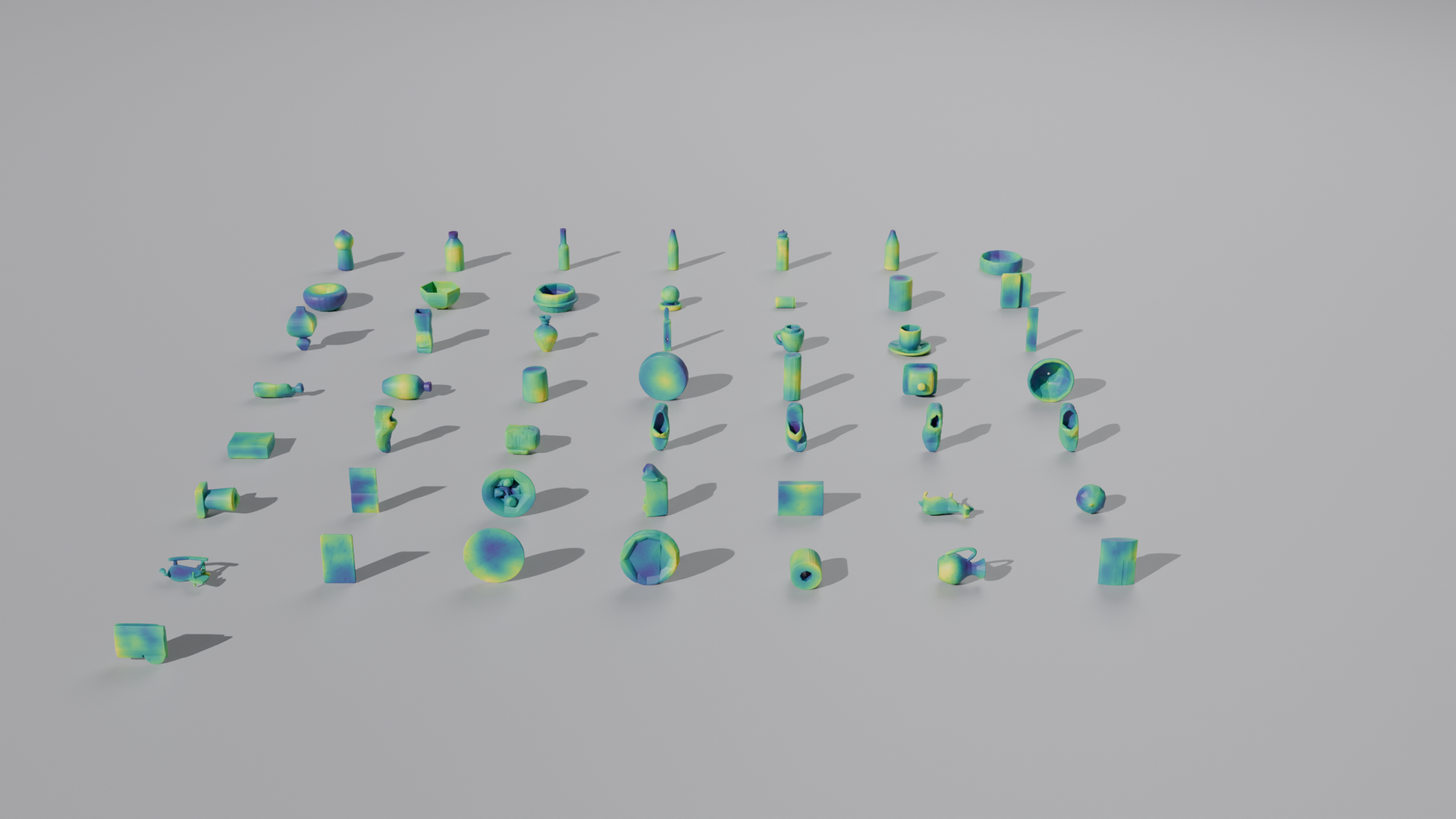}
        \caption{Allegro - default}
        %\label{fig:image1}
        \vspace{0.5cm}  % space between rows
    \end{subfigure}
    \begin{subfigure}[b]{0.49\textwidth}        
    \includegraphics[trim={4cm 6cm 7cm 8cm},clip, width=\linewidth]{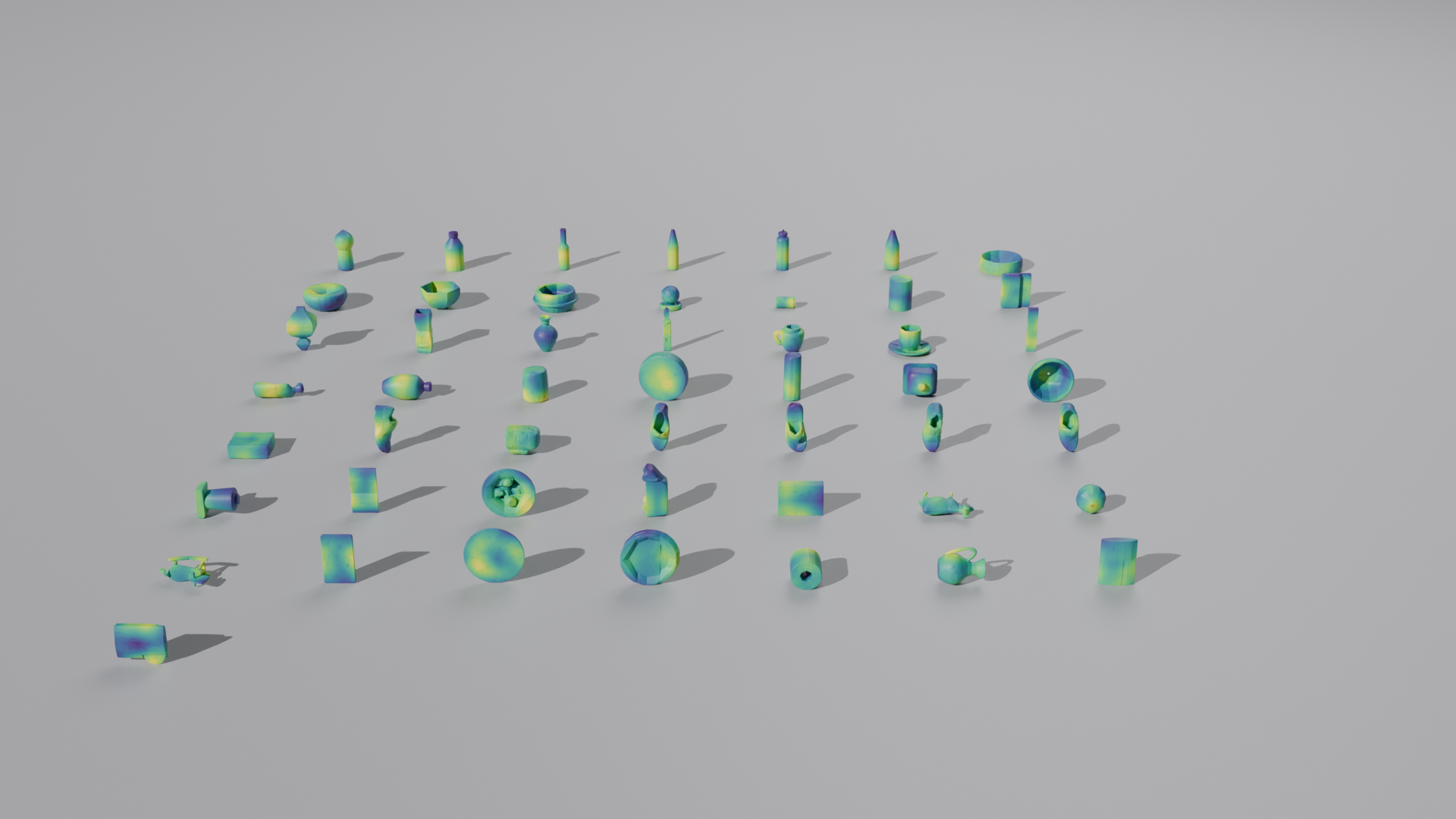}
        \caption{Allegro - precision}
        %\label{fig:image2}
    \end{subfigure}
    \hfill
    \begin{subfigure}[b]{0.49\textwidth}        
    \includegraphics[trim={4cm 6cm 7cm 8cm},clip, width=\linewidth]{latex_figures/allegro_12_contacts_span_overall_cone_sqp_default_longer_gendex_precision_render.png}
        \caption{Allegro - pinch}
        %\label{fig:image2}
    \end{subfigure}
    \caption{
    Contact heatmaps for various objects using the Allegro hand. The color intensity (mapped with the viridis colormap) indicates regions of frequent contact across all grasp poses: yellow represents high-contact areas, while blue indicates low-contact regions. Notably, many objects show concentrated contact regions near their center of gravity.
    }
\end{figure}

\end{document}